\documentclass[letterpaper, 10 pt, conference]{ieeeconf}  % Comment this line out if you need a4paper

\IEEEoverridecommandlockouts                              % This command is only needed if 
\usepackage{graphicx} % for pdf, bitmapped graphics files
\usepackage[english]{babel}
\usepackage[utf8x]{inputenc}
\usepackage{graphicx}
\usepackage{amsfonts}
\usepackage{flushend}

\usepackage{amsmath}
\usepackage{amssymb}
\usepackage{array}
\usepackage{xcolor}
\usepackage[ruled,vlined]{algorithm2e}
\usepackage{subcaption}
\usepackage{multicol}
\usepackage{hyperref}
\usepackage{url}
\usepackage{footmisc}
\usepackage{siunitx}

\newcommand{\cA}{\mathcal{A}}
\newcommand{\cS}{\mathcal{S}}
\newcommand{\cO}{\mathcal{O}}

\newcommand{\cN}{\mathcal{N}}

\newcommand{\cP}{\mathcal{P}}
\newcommand{\cR}{\mathcal{R}}
\newcommand{\cD}{\mathcal{D}}
\newcommand{\cI}{\mathcal{I}}

\newcommand{\R}{\mathbb{R}}

\newcommand{\E}{\mathbb{E}}

\newcommand{\tnorm}[1]{\left\lVert#1\right\rVert_2}

\newcommand\numberthis{\addtocounter{equation}{1}\tag{\theequation}}

\newcommand{\bx}{\mathbf{x}}
\newcommand{\bX}{\mathbf{X}}

\newcommand{\bp}{\mathbf{p}}
\newcommand{\bv}{\mathbf{v}}
\newcommand{\bt}{\mathbf{t}}

\newcommand{\bZ}{\mathbf{Z}}
\newcommand{\bK}{\mathbf{K}}

\newcommand{\td}{\dot{\theta}}

\newcommand{\st}{^\star}
\newcommand{\sst}{^{\star\star}}

\newcommand*\diff{\mathop{}\!\mathrm{d}}

\newcommand{\Var}{\mathrm{Var}}
\newcommand{\Cvar}{\mathrm{CVaR}}

\renewcommand{\epsilon}{\varepsilon}
\newcommand{\eps}{\epsilon}

\DeclareMathOperator*{\argmin}{arg\,min}
\DeclareMathOperator{\dis}{\Gamma}
\DeclareMathOperator*{\sign}{sign}

\newcommand{\trsp}[1]{#1^\intercal}
\newcommand{\inv}[1]{#1^{-1}}

\usepackage[]{algorithm2e}
\usepackage{ragged2e}
\usepackage{etoolbox}
\usepackage{empheq}

\title{\LARGE \bf
Safe Sampling-Based Air-Ground Rendezvous Algorithm for Complex Urban Environments*
}

\author{Gabriel Barsi Haberfeld$^{1}$, Aditya Gahlawat$^{2}$, and Naira Hovakimyan$^{3}$% <-this % stops a space
\thanks{*This work was supported by NSF NRI award 1830639.}% <-this % stops a space
\thanks{$^{1}$Gabriel Barsi Haberfeld is a Ph.D. Student with the Department of Mechanical Science and Engineering, University of Illinois at Urbana-Champaign
{\tt\small gbh2@illinois.edu}.}%
\thanks{$^{2}$Aditya Gahlawat is a Postdoctoral Researcher with the Department of Mechanical Science and Engineering, University of Illinois at Urbana-Champaign
{\tt\small gahlawat@illinois.edu}.}%
\thanks{$^{3}$Naira Hovakimyan is with Faculty at the Department of Mechanical Science and Engineering, University of Illinois at Urbana-Champaign
{\tt\small nhovakim@illinois.edu}.}%
}

\begin{document}

\maketitle
\thispagestyle{empty}
\pagestyle{empty}

%%%%%%%%%%%%%%%%%%%%%%%%%%%%%%%%%%%%%%%%%%%%%%%%%%%%%%%%%%%%%%%%%%%%%%%%%%%%%%%%
\begin{abstract}

Demand for fast and economical parcel deliveries in urban environments has risen considerably in recent years. A framework envisions efficient last-mile delivery in urban environments by leveraging a network of ride-sharing vehicles, where Unmanned Aerial Systems (UASs) drop packages on said vehicles, which then cover the majority of the distance before final aerial delivery. Notably, we consider the problem of planning a rendezvous path for the UAS to reach a human driver, who may choose between $N$ possible paths and has uncertain behavior, while meeting strict safety constraints. The long planning horizon and safety constraints require robust heuristics that combine learning and optimal control using Gaussian Process Regression, sampling-based optimization, and Model Predictive Control. The resulting algorithm is computationally efficient and shown to be effective in a variety of qualitative scenarios.

\end{abstract}

%%%%%%%%%%%%%%%%%%%%%%%%%%%%%%%%%%%%%%%%%%%%%%%%%%%%%%%%%%%%%%%%%%%%%%%%%%%%%%%%
\section{INTRODUCTION}
\label{sec:intro}

\begin{figure}
        \centering
        \begin{subfigure}[b]{0.26325\textwidth}
            \centering
            \includegraphics[width=\textwidth]{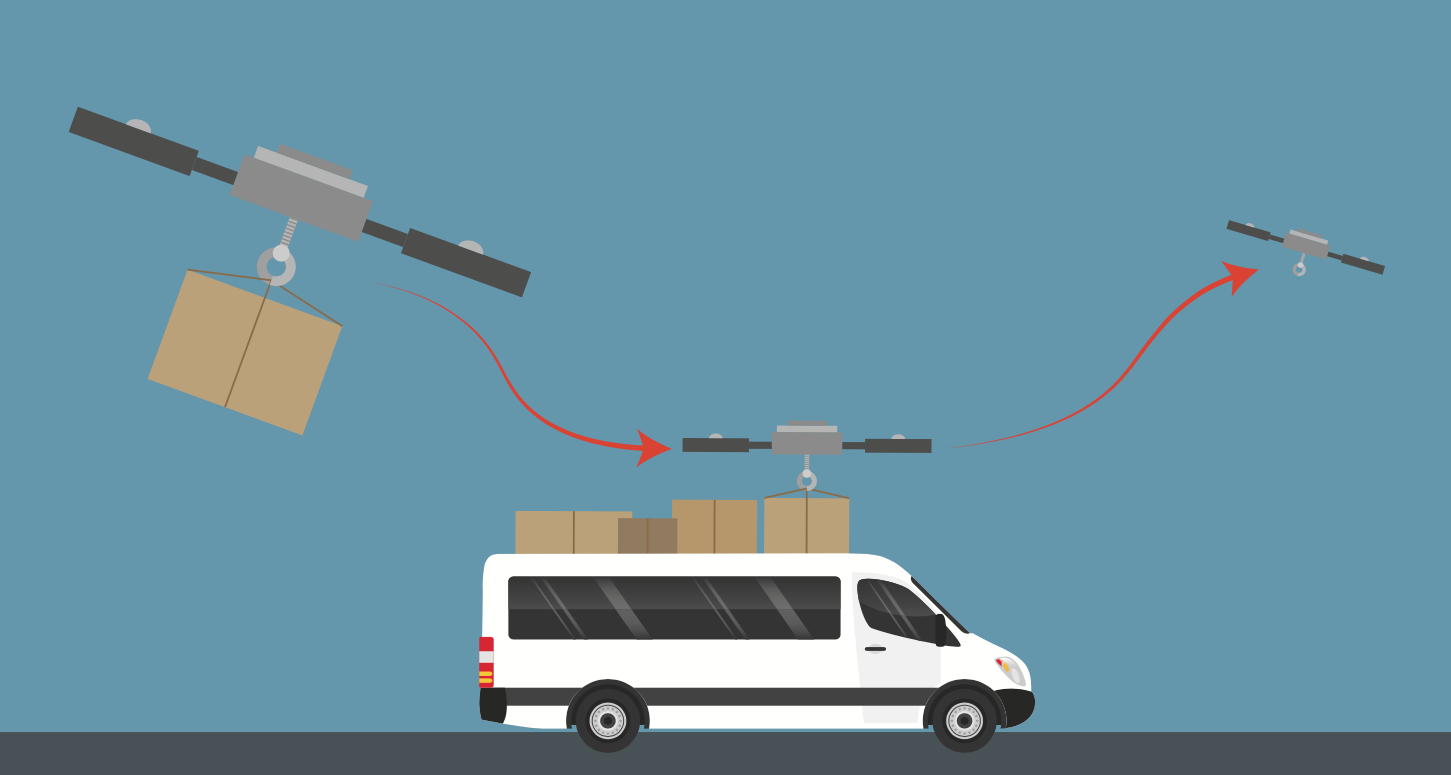}
            \caption[]%
            {{}}    
            \label{fig:proc}
        \end{subfigure}
        \begin{subfigure}[b]{0.16875\textwidth}  
            \centering 
            \includegraphics[width=\textwidth]{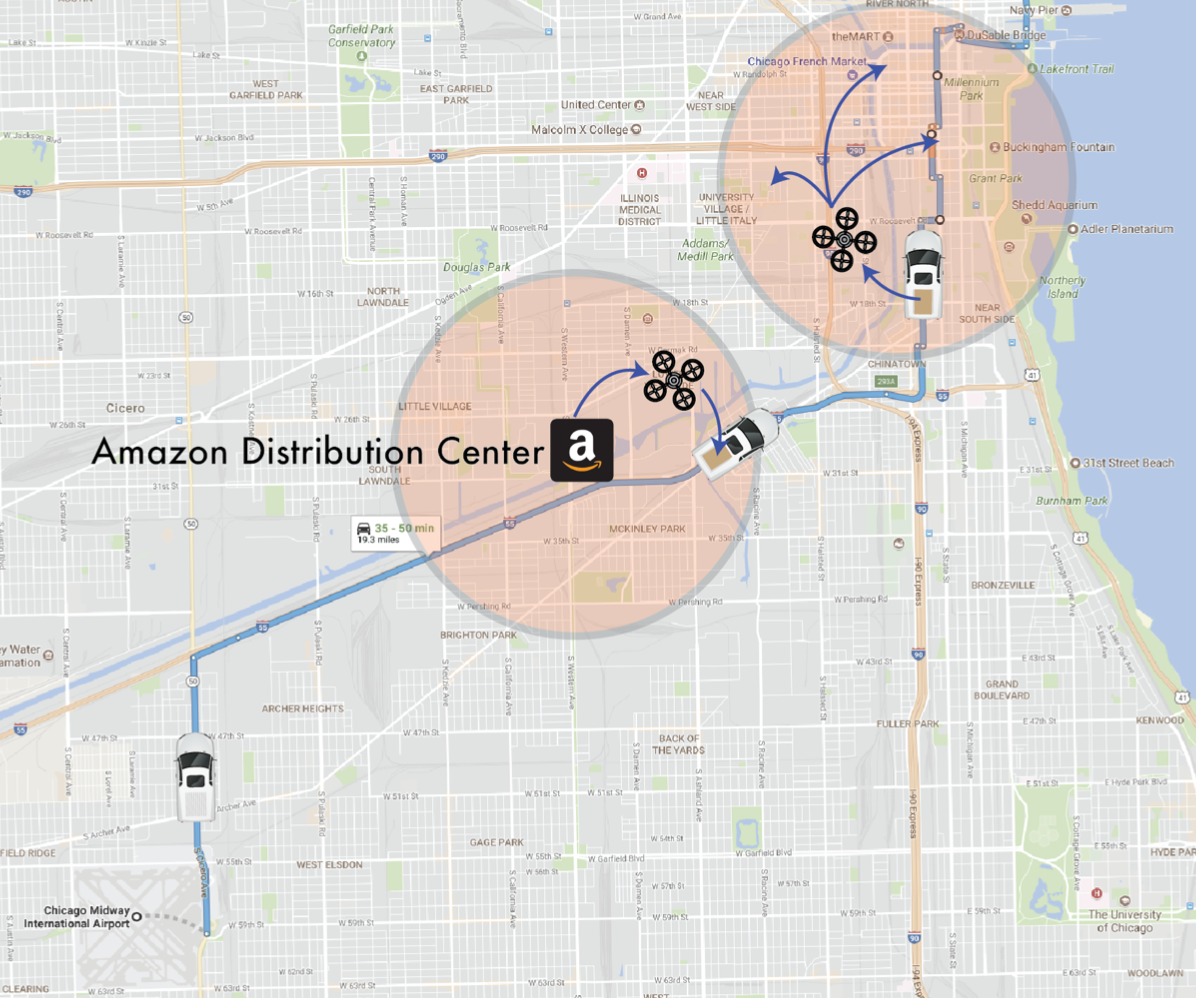}
            \caption[]%
            {{}}    
            \label{fig:longh}
        \end{subfigure}
        \caption[]
        {\small Air-ground rendezvous procedure. Left: the UAS needs to meet an uncontrollable ground vehicle with uncertain trajectory. Right: UASs intercept the vehicle at various points to complete the delivery. In this example an Amazon package is carried by a ride-sharing vehicle departing from Chicago Midway Airport bound to Downtown Chicago.} 
        \label{fig:introfig}
\end{figure}

Modern transportation solutions can accumulate more than half of the total shipping cost on the transportation portion between the final distribution center and the customer \cite{mckinsey}. This fact is known as the {\em last-mile problem}. Our proposed framework consists of using the existing large networks of ride-sharing services (Uber, Lyft) to cover most of the distance from the final distribution center to the customer. This process uses knowledge of these vehicles' destination to plan deliveries, where a UAS carries the parcel from the distribution center and places it on a moving vehicle or picks up a package from a moving vehicle and delivers it to a customer. An example scenario is illustrated in Fig.~\ref{fig:introfig}. The critical concern is driver behavior. An erratic driver adds an undesirable risk to the two stages of the mission: (1) landing safely on the moving vehicle to drop the parcel and (2) flying back to the distribution center. Environmental factors such as wind, package mass, sloshing of package contents, battery age, and others contribute to these safety concerns. However, because of the long planning horizons associated with these missions, the primary source of risk and uncertainty arises from the inaccurate driver behavior, where a driver might be slower, faster, or generally unpredictable. In this paper, we build on our previous solution for simple missions \cite{acc-risk}. The main extension is that the method now admits non-differentiable paths and allows the uncertainty of path choice, where a driver might choose a different route than the one shared with the UAS a priori.

Model Predictive Control (MPC) is a  popular method for solving local Optimal Control Problems (OCP) in real-time \cite{mesbah}, where the OCP is solved at each iteration of the control loop. Although versatile, traditional MPC is not equipped to deal with large uncertainties over long planning horizons due to exponentially increasing uncertainty propagation in the planning stage. To address these issues two common solutions are (a) stochastic MPC (SMPC) \cite{stochasticmpc,primbs} and (b) Robust MPC \cite{robustmpc,park}. Stochastic MPC is often referred to as \emph{risk-neutral}, as it aims to minimize expectations singularly, while Robust MPC accounts for worst-case scenarios. In some cases, an absolute approach is desirable, but often the problem requires a trade-off between high risk and robustness, as not to diverge too far away from optimality. In the context of urban aerial logistics, we aim to minimize the risk of running out of battery and inevitably crashing. \textcolor{black}{In this paper, we handle tractability problems of optimizing over risk measures \cite{Schildbach_2014} with a gradient-free sampling-based approach. The method shown in Section \ref{sec:meth} also allows planning over multiple non-differentiable paths.}

%In most cases, optimizing over risk measures turns the OCP intractable for real-time implementation due to the added complexity \cite{Schildbach_2014}. Additionally, gradient-based approaches will underperform or entirely fail if there are multiple path possibilities or non-differentiable paths. In this work, we use an MPC-like algorithm to find optimal risk-averse trajectories and tackle the deficiencies mentioned earlier with a combination of stochastic optimization and probabilistic heuristics as exposed in Section \ref{sec:meth}.

In our previous work \cite{acc-risk} we consider the task planning problem of guiding a UAV to the neighborhood of a human-operated vehicle traveling along a known path. Uncertain driver behavior and the large distances the UAS needs to cover required the parallel planning of one risk-enabled path that rendezvous with the driver and one deterministic return path. The motivation is that under the conditions where a rendezvous is only safe if there is a low probability of running out of battery, we can increase the potential range of the mission if we have high assurance that the UAS can meet the driver, and hence have less payload on the way back. Should we commit to such a plan with a high risk of not meeting the driver, there is a high probability that the package will not be delivered, and the extra mass will cause the UAS to deplete its battery prematurely and crash on the way back. 

%Figure \ref{fig:ranges} depicts the potential range increase.
%\begin{figure}
%    \centering
%    \includegraphics[width=0.45\textwidth,trim=90mm 60mm 180mm 0mm, clip]{ACC-Risk-2020/fig/ranges.pdf}
%    \caption{Different maximum ranges for different assumptions. A plan that assumes no payload is futile, and one that assumes a payload the entire time is conservative. Given an assurance that the payload will be dropped half-way through the mission, we can safely plan on the red region, increasing range over the green region.}
%    \label{fig:ranges}
%\end{figure}
In this work, we allow the driver not to be constrained to a single path; instead, the driver can choose among $N$ different parametrized paths. 
%The naive approach of computing a single-path solution for each path and tracking the one with the least risk is unfit for two reasons. First, planning for a single path is overly ambitious as there is an unacceptable probability that another path will be chosen, causing an abort decision or an accident. Second, as $N$ increases, the computational load will be unbearable using the traditional gradient-based nonlinear solver. 
We propose a sampling-based method similar to that in \cite{CEMotion}. After selecting the best sample as a rendezvous location candidate, an MPC-like controller generates inputs for two trajectories. One trajectory rendezvous with the car in future time and another returns to a safe landing location. The decision between which trajectory to follow is made by probabilistic heuristics that monitor risk measures based on the sampling statistics. In Section \ref{sec:problem} we formalize the problem setup.
%
%The control algorithm loop follows the steps:
%\begin{enumerate}
%    \item Build a driver's model using Gaussian-Process Regression \cite{williams2006gaussian} from local sensor data.
%    \item Sample possible rendezvous locations for each path from an $N$-dimensional Gaussian distribution $\mathcal{A}$.
%    \item Assign the top $n_e$ samples according to a cost function $l$ to a set $\mathcal{S}_e$
%    \item Update $\mathcal{A}$ with the statistics from $\mathcal{S}_e$.
%    \item Plan and incrementally execute a rendezvous trajectory resulting from an Optimal Control Problem (OCP).
%    \item Plan a deterministic abort trajectory that returns the UAS to a safe landing location.
%\end{enumerate}

\subsection{Related work}
Several papers have considered risk measures in planning and handling uncertainties in an MPC framework, as summarized in \cite{mesbah,mayne}, and shown in \cite{park,cautiousmpc,pavone,riskbook,riskmeasures,whittle,Sopasakis}. In \cite{cautiousmpc}, the authors study uncertainty propagation to ensure chance constraints on a race car; results show that the algorithm can learn uncertainty in the dynamics, associating risk with the unknown dynamics, and plan so that the trajectories are safe. In \cite{pavone}, the authors provide stability proofs for a linear MPC controller, which minimizes time-consistent risk metrics in a convex optimization form. These papers focus on operating in a constrained environment or under controlled assumptions to provide uniform guarantees. Our work's key difference is that we relinquish online risk constraint satisfaction to external heuristics, widening the solver's capabilities and flexibility at the cost of a more conservative solution. Apart from fundamental results in this field, such as \cite{whittle}, modern developments in \cite{Sopasakis} show that the increased computational capacity enables executing risk-minimization in real-time for a variety of systems. Few papers have been published concerning highly stochastic rendezvous problems. Most notably, in \cite{rucco} the authors compute optimal trajectories in refueling missions, but in their work, most of the uncertainty is environmental and local, whereas we consider epistemic and large-scale uncertainties. %Risk minimization in optimal control has been studied in several papers \cite{cautiousmpc,pavone,whittle,Sopasakis,fleming}. Minimizing risk in optimal control is traditionally intractable  for real-time implementation \cite{cautiousmpc,pavone}, a capability which the proposed method is designed to have.

Lastly, sampling-based motion planning has grown in popularity with the increased computational power afforded by modern CPUs and Graphics Processing Units (GPUs). The core idea behind these methods is to cleverly sample inputs from some distributions and use these samples' quality to update the distribution and improve the next batch of samples. Ultimately, the goal is to converge to a narrow distribution centered around the optimal input (or set of inputs) to the system. In this paper, we sample rendezvous times and their associated rendezvous location. Two approaches are directly related to this paper. In \cite{CEMotion} the authors present a novel method to find trajectories for mobile robots in cluttered environments. In \cite{evangcvar}, the authors present a sampling-based MPC that integrates risk management using Conditional Value-at-Risk ($\Cvar$) ~\cite[Sec.~3.3]{riskbook}. Both motion planning and usage of $\Cvar$ are relevant to this paper, both of which are described in Section \ref{sec:meth}.

\subsection{Problem Novelty}
\textcolor{black}{The rendezvous (or interception) problem is not new. Traditionally these problems fall into two categories: interception of a target on a known path or interception of a target with an unknown trajectory \cite{pursuitevasion}. Full knowledge of target behavior makes the problem trivial as shown in \cite{ethtraj}. In these scenarios, the goal is to find the \textit{optimal} trajectory for interception. Conversely, no knowledge of target behavior requires problem relaxation \cite{robotlos,missle} with little guarantees. In this paper, we require certain constraints usually not afforded by latter, and have only partial target behavior knowledge. Thus, we require a custom solution that exploits the particulars of the rendezvous problem to provide safety guarantees.}

\subsection{Statement of contributions}

We present a hybrid algorithm that is capable of attempting to rendezvous with a human-operated ground vehicle and does not crash with guaranteed safety bounds. The algorithm handles three significant challenges: driver behavior, multiple possible routes, and non-differentiable paths. To safely attempt a rendezvous, three main components are needed.

A Gaussian Process Regression learning module collects sensor data from the ground vehicle and builds a driver model. Unique to this paper is the way we pose this learning problem. Instead of modeling future driver position, we leverage historical traffic data from the area to learn how the driver deviates from an average virtual driver. This way, we significantly reduce the problems associated with uncertainty propagation. This benefit is only possible by the fact that we know all reachable roads a priori.

To address pathing complexities, we use a modified version of \cite{CEMotion} that accepts a specialized risk measure and integrates with the learning and MPC layers instead of finding optimal trajectories directly. This novel way of sampling trajectories relies on fast Gauss-Kronrod Quadratures to estimate sampling quality, an approach that can benefit from parallel computing hardware such as GPUs.

The last component is an MPC-like controller. Unlike traditional MPC, our formulation uses the time horizon as an input, allowing a compact set of variables capable of timing control. The temporal component is crucial because an optimal rendezvous has the UAS reaching the ground vehicle precisely both in space and time. To achieve this, we use the fact that the mission's spatial scale is large enough that single-integrator dynamics are a satisfactory approximation. 

These three modules share critical information backward and forwards between each other and provide a robust, efficient, and concise set of hyperparameters. The rest of this paper is structured as follows: in Section \ref{sec:problem}, we introduce and define the problem in algorithmic format. In Section \ref{sec:meth}, we present the three main components of this approach: the model learning, the importance sampling scheme, and the MPC controller. In section \ref{sec:res}, we demonstrate results for individual modules and two full planning stack examples. Finally, in Section~\ref{sec:conc}, we provide concluding remarks and discuss the shortfalls of this approach and future directions to address them, respectively.

\section{Problem Formulation}
\label{sec:problem}

We begin by defining the notion of persistent safety.
\newtheorem{definition}{Definition}
\begin{definition}[Persistent Safety]
\label{def:persafe}Let $x_{k+1}=f(x_k,u_k)$ be a dynamical system with state vector $x\in\R^n$ and control vector $u\in\R^m$. A safety set $S_k\subset(X,U)$ is a set, in which all states and inputs are considered safe by some measure $\rho(x):\R^n\rightarrow\R$ at step $k$. We define a planning algorithm as persistently safe, if $S_k = \{x_k\in X,\;u_k \in U : f(x_k,u_k) \in S_{k+1}\}$ exists for all $k$ for a set of admissible states $X$ and control inputs $U$.
\end{definition}

The goal is to compute a persistently safe trajectory (sequence of states $x$ and inputs $u$) as defined in Definition \ref{def:persafe} that satisfies a rendezvous condition. This computation is achieved by postponing a decision between aborting or continuing the mission for as long as possible. The additional time afforded by postponing this decision is used to improve uncertainty prediction and, consequently, reducing the risk of running out of battery or fuel.

Consider a set $\cP$ of $N$ indexed parametrized paths $\cP=\{p_j(\theta),\;j=1,\dots,N\}\subset\cR$, $p:\R^+\rightarrow \mathbb{R}^2,\;\theta\in\R^+$ on a planar Euclidean region $\cR\in\R^2$, and historical velocity data along each path $\dot{\theta}^h_j(t),\;\td^h_j:\R^+\rightarrow\R$ obtained from the traffic data that are provided apriori. A stream of noisy position $\theta^d(t)$, $\theta^d:\R^+\rightarrow\R^+$, and velocity $\td^d_j(t)$ measurements from a driver moving along  any of the paths are obtained in real-time via on-board sensors. Let $\varkappa$ be set of all permutations of $\{1,\dots,N\}$. Let a path intersection set be defined as $\cI=\{\{\theta,p\}\in\R^2|p_i(\theta)=p_j(\theta),\{i,j\}\in\varkappa\}$. The intersection set $\cI$ contains all path segments that intersect each other before terminally pruning. The purpose of this set is to identify values of $\theta$ for which we are uncertain of what path the driver will choose. We ignore paths that the driver may no longer choose, i.e. a path which the driver passed and chose not to turn into. Obviously this also makes it so that, within $\cI$, the historical data is the same across intersecting paths. We wish to find a rendezvous point $\theta^d_j(t_{R,j})$ that brings both vehicles together at a rendezvous time $t_{R,j}\in\R^+$, for some path $j$. 

Due to sensor noise and uncertain driver behavior, we aim to learn the distribution of $\theta^d_j(t_{R,j})$ and plan on it. In single path problems the distribution $g(\theta_d|t_R)$ is distributed along the path. For this problem, however, each path will have its own distribution $g_j(\theta^d_j|t_{R,j})$. We approximate driver behavior by learning a deviation mean function $d(\theta^h_j(t)):\R\mapsto\R$ and variance function $\Sigma_d(\theta^h_j(t)):\R\mapsto\R$. The deviation function is such that $\theta^d_j(t)=\theta^h_j(t)+d(\theta^h_j(t))$ if learned exactly. Section \ref{sec:gpd} explains this learning process.

%so that $\cT =\{g_j(\theta^d_j|t_{R,j}),\;j=1,\dots,N\}$ is an $N$-dimensional distribution. For the purposes of this work, we consider $\cA = \cN^N(\mu,\Sigma)$ where $\cN^N$ is an $N$-dimensional Gaussian distribution with mean function $\mu_d:\R\mapsto\R$ and covariance function $\Sigma_d:\R\mapsto\R$. 
The next step is to use the driver model to find quality rendezvous candidates comprised of time and location pairs. This search is done through stochastic optimization, explained in Section \ref{sec:ce}. Assume each path has an optimal rendezvous location local to the path. The random nature of the problem makes the rendezvous locations stochastic, with an $N$-dimensional distribution $\cA\st(\mu_\cA\st,\Sigma_\cA\st)$. Since we do not have knowledge of $\cA\st$ we aim to approximate it by manipulating the parameters of an ancillary distribution $\cA(\mu_\cA,\Sigma_\cA)$ with equal dimension. To estimate $\cA$ we require a driver model ($d$ and $\Sigma_d$) and some way of optimizing the parameters of $\cA$ such that $\cA\rightarrow\cA\st$ as $t\rightarrow\infty$. If we are successful, we can find the optimal path to rendezvous with defined as $p_\mathrm{tgt}$, such that $p\st = p_\mathrm{tgt}(\theta^d_\mathrm{tgt}(t_{R,\mathrm{tgt}}))$ is the optimal rendezvous location.

\begin{figure}
    \centering
    \includegraphics[width=0.35\textwidth,trim=190mm 100mm 200mm 20mm, clip]{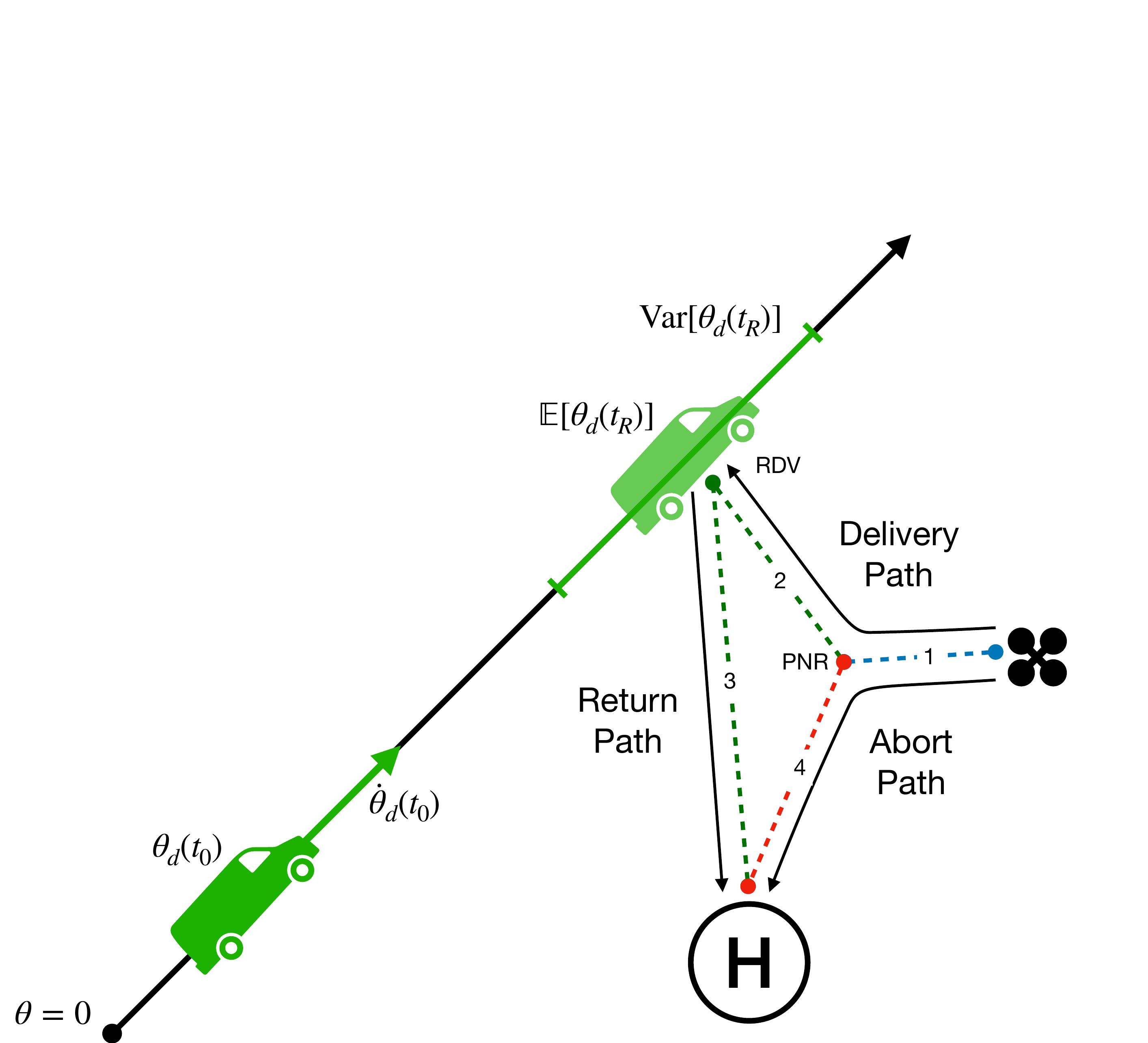}
    \caption{Overview of the problem setup at time instance $t_0$ for a single path. Uncertainty in driver behavior and path choice requires multiple plans. Each path has associated risk and cost. We plan a Point-of-No-Return to afford extra time for data acquisition.}
    \label{fig:overview}
\end{figure}

\begin{figure}
    \centering
    \includegraphics[width=0.35\textwidth,trim=170mm 80mm 160mm 50mm, clip]{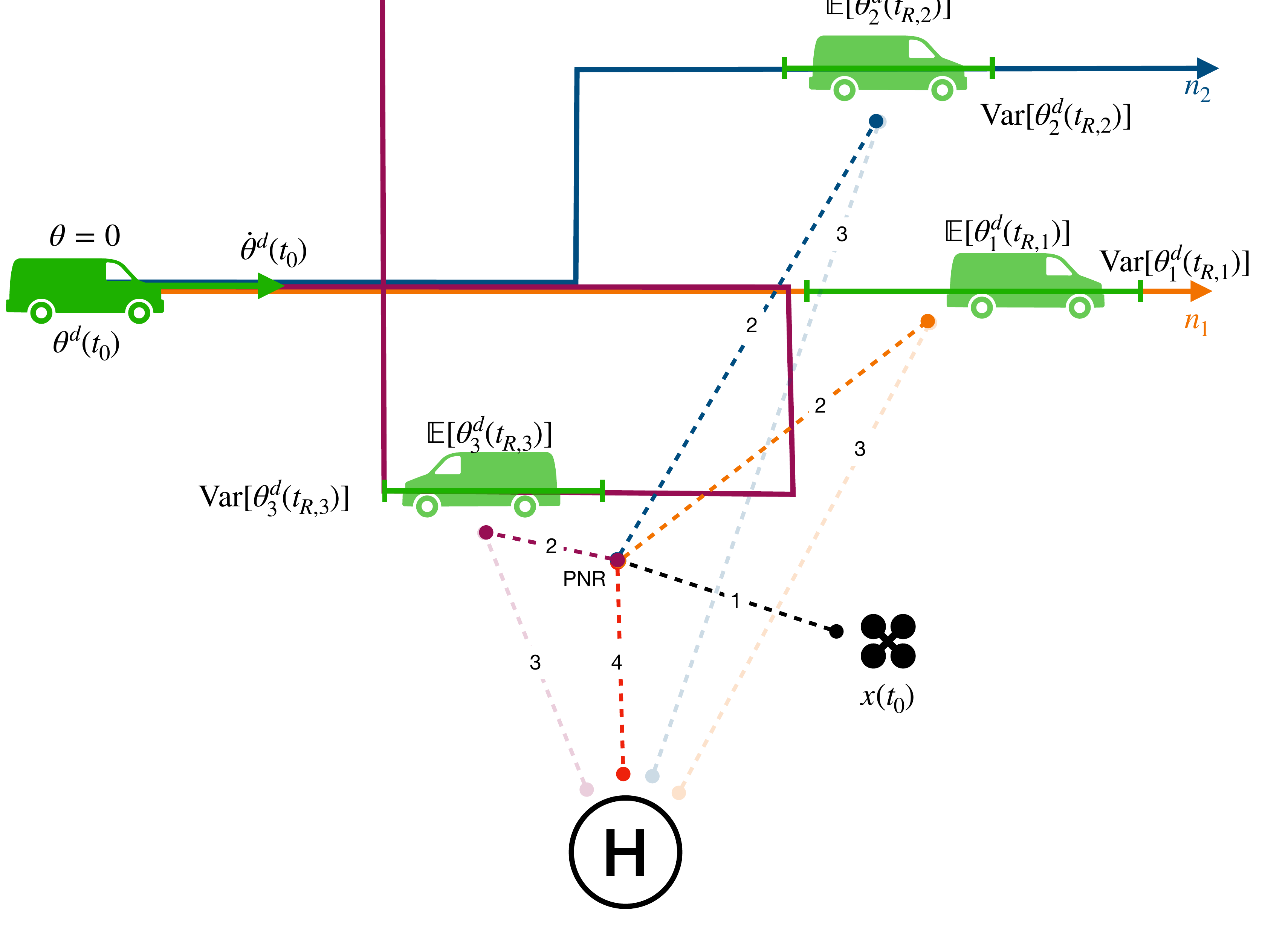}
    \caption{Overview of the problem setup at time instance $t_0$ for  multiple paths. Additional uncertainty in path choice: each path has it's own individual driver position uncertainty. Planning for all outcomes is intractable.}
    \label{fig:overview-mp}
\end{figure}

With knowledge of $p\st$, the next step is to find a trajectory that guides the UAS to that point in space and time. We wish to have guarantees that the UAS will not run out of battery. To achieve this, the trajectory planner plans two options; one that rendezvous with the ground vehicle and another that returns to a safe landing location. Because the latter (abort path) does not depend on any uncertainties, we are guaranteed to land safely by choosing that option. The cost is that we do not rendezvous and render the system sub-optimal. To mitigate this problem, we find the two paths by planning a Point-of-No-Return (PNR) between the UAS and $p\st$, from which a separate path navigates the UAS to a safe landing location in case the risk of rendezvous failure $\rho_d(p\st)$ is too great. The risk measure $\rho_d$ maps the distribution of $p\st$ and system states to $\R^+$. We model the UAS as a system capable of tracking single integrator dynamics in this context. We define safety (and, thus, its associated risks) as a function of the probability of running out of remaining battery or fuel $E_r$. Figure \ref{fig:overview} illustrates the setup for a single path. In Section \ref{sec:almain} we present the multi-path setup, which is illustrated in Figure \ref{fig:overview-mp}.

The discretized single integrator dynamics of the UAS are
\begin{align}
    x_k &= x_{k-1} + v_kT_s,\label{eq:sysdyn}
\end{align}
where $x_k\in\R^2$ is the Euclidean UAS position, $v_k\in\R^2$ is the Euclidean velocity input, and $T_s$ the sampling time. Additionally, the remaining energy has the following dynamics:
\begin{align}
    E_{r,k} &= E_{r,k-1} - \left(\dfrac{mv^2}{2} + \alpha m\right) T_s,\label{eq:edyn}
\end{align}
where $E_{r,k}$ is the remaining energy, $m\in[m_a,m_b],\;m_a>m_b$ is the mass of the UAS (for a package drop-off mission, the mass decreases after the package is dropped on the ground vehicle), and $\alpha$ is the scalar hovering energy consumption constant. %We denote $x(t)$ and $E_r(t)$ as the continuous-time realization of \eqref{eq:edyn} and \eqref{eq:edyn} respectively, where $x_k=x(k\cdot T_s)$. 
Note that $m$ will decrease from $m_0$ to $m_1$ after the package is dropped off on the ground vehicle, decreasing the energy consumption rate and increasing the range. This is the detail that makes it beneficial to commit to plan so that we can reach further and increase efficiency. Additionally, given this problem's large scale, the single integrator assumption is not strong. At this scale virtually any controllable robot will be able to track these dynamics without difficulties. Adaptive control techniques such as \cite{pereida2018adaptive,cao2008design} can formalise this assumption. We can now present the problem formulation. 
\newtheorem{formulation}{Problem}
\begin{formulation}[Risk-Averse Multi-Path Rendezvous]
\label{problem}
Given a map of $N$ paths, historical velocity data along each path $\dot{\theta}^h_i(t)$, and a stream of noisy position $\theta^d(t)$ and velocity $\td^d(t)$ measurements from a driver traversing an intersection of any subset of the paths, find a persistently safe sequence of control inputs and a future time $t_R$ such that the UAS reaches a neighbourhood of the driver at time $t_R$ and flies to a safe pre-determined landing location. This trajectory is the solution to the following optimization problem
\begin{alignat*}{3}
    \min_{U,t_R,t_L}\quad   & L(x_k,u_k,\cD) - t_D & \\
    \text{s.t.}\quad    & x_k = x_{k-1} + v_kT_s &\\
    & ||x(t_{R,\mathrm{tgt}})-p\st||\leq \eps,\;x(t_L)=S_L &\\
    & E_{r,k} = E_{r,k-1} - \left(\dfrac{mv^2}{2} + \alpha m\right) T_s &\\
    & E_{r}(t_{R\st})\geq 0,
\end{alignat*}
where $L(\cdot)$ is a cost function that minimizes risk, input costs, and delivery time, $t_D$ is the decision time between the current state $x(t_0)$ and Point-of-No-Return (PNR), $t_{R,\mathrm{tgt}}$ is a rendezvous time for path $\mathrm{tgt}$, $p\st$ is a rendezvous location, $\eps$ is a small positive number, $t_L$ is a landing time for a landing location $S_L\in\R^2$, $x(t)$ and $E_r(t)$ are continuous time realizations of $x_k$ and $E_{r,k}$ respectively, and $\mathcal{D} = \{D,H\}$ is a data set containing measurements from the ground vehicle, where $D,~H \in \mathbb{R}^M$ are defined as
\[
D = \begin{bmatrix} \td^{d,1} & \cdots & \td^{d,M} \end{bmatrix}^\top,     \quad 
H = \begin{bmatrix} \td^{h,1} & \cdots & \td^{h,M} \end{bmatrix}^\top,
\]
and $M$ denotes the number of measurements collected, $\td^{d,j}$ are the driver samples, and $\td^{h,j}$ is the expected velocity at the GPS-collected point $\theta^{d,j}$ obtained from historical data.
\end{formulation}

In Section \ref{sec:meth} we present the tools that solve Problem \ref{problem} by altering its different components. Although we never solve the Optimal Control Problem shown above explicitly, we reach an equivalent solution through multiple planning stages.

\section{Methods}
\label{sec:meth}

\subsection{Driver Model Learning}
\label{sec:gpd}
This section discusses the learning component introduced in Section \ref{sec:problem}. One of the major challenges for the proposed problem is that each driver behaves differently. While one driver may drive at a conservative speed limit, another might drive relatively faster, slower, or erratically. Therefore, learning a driver's `behavior' will be beneficial to the rendezvous problem. Later on, we use this model in the approximation algorithm that estimates future driver position. We now set up this learning problem. We assume that we have access to the driver's position $\theta^{d,i} = \theta^d(t^i)$, where $t^i$ is the time instance at which the measurement is obtained. Furthermore, we have measurements of the driver's velocity denoted by $\dot{\theta}^{d,i} = \dot{\theta}^d(t^i)$. Note that there is no dependency on which path the driver is on because we assume that $\theta^{d,i}\in\cI$, and $\cI$ ignores non-reachable paths. All measurements are considered to have additive normally distributed noise. We also assume that we have access to historical velocity profile given by $\dot{\theta}^{h,i} = \dot{\theta}^h(t_i)$. Such a historical velocity profile can be generated by collecting measurements of vehicles traversing the path and fitting a distribution over it using methods similar to those in \cite{traffic1,traffic2}. In our case, we assume the historical velocity profiles are in the form of time-parametrized mean functions. To summarize, given the driver's position $\theta^{d,i}$, we have access to a measurement of the driver's velocity $\dot{\theta}^{d,i}$ and the corresponding probabilistic historical velocity $\dot{\theta}^{h,i}$. A comparison of $\dot{\theta}^{d,i}$ and $\dot{\theta}^{h,i}$ thus represents a measure of the driver's behavior. In particular, we wish to learn $\dot{\theta}^d(\dot{\theta}^h): \mathbb{R} \rightarrow \mathbb{R}$. An equivalent problem is to learn a deviation function $d(\dot{\theta}^h): \mathbb{R} \rightarrow \mathbb{R}$ such that $\dot{\theta}^d(t) = \dot{\theta}^h(t) + d(\dot{\theta}^h(t))$. Throughout this paper, we learn the deviation function.
%
%Figure \ref{fig:fitting} shows an example where we learned an approximation of $d$ using Gaussian Process Regression.
%   \begin{figure}
%      \centering
%      \includegraphics[trim=0mm 0mm 0mm 0mm, clip, width=0.45\textwidth]{ACC-Risk-2020/fig/gp_perf.png}
%      \caption{Example regression: a nonlinear curve we wish to learn (red dashed) maps historical data to driver behavior, by regressing on the measurements (dots) we get: mean (solid line) and variance functions (blue shaded area). Away from the data points prediction becomes unreliable.}
%      \label{fig:fitting}
%   \end{figure}

The traditional approach would be to directly learn the vehicle's position function $\theta^d(t)$; however, this would cause the uncertainty propagation to expand too quickly and force an abort decision too often \cite{cautiousmpc}. Instead, we explore both the fact that the vehicle is constrained to a known path and that the velocity along the path has a strong prior (the historical velocity $\dot{\theta}^h(\cdot)$). A disadvantage of this approach is that an integration procedure must be carried out to estimate $\theta^d(t)$. 
%In a regular OCP formulation, this function would be integrated inside the solver as dynamic model constraints \cite{borrelli}. However, due to the coarse discretization considered in this paper, such implementation would not be feasible. 
\textcolor{black}{We leverage the sampling-based nature of this algorithm presented in Subsection \ref{sec:ce} and modern numerical integration methods to provide a computationally efficient integration procedure.}

As described by Williams and Rasmussen \cite{williams2006gaussian}, a Gaussian Process is a generalization to functions of the Gaussian distribution. Assume that we have a stream of $M\in\mathbb{N}$ measurements of the form $y_i = d(x_i) + \zeta$, $\zeta \sim \mathcal{N}(0,\sigma_n^2)$, $i \in \{1,\dots,N\}$, where $y_i=x_i-\td^{d,i}$, $\td^{d,i}$ is the actual sensor measurement, and $x_i=\td^{h,i}$ is known a priori. Note that this definition is equivalent as far as learning objectives to the one in Problem \ref{problem}. Now let $\mathbf{Y} = \trsp{\begin{bmatrix} y_1 & \dots & y_M \end{bmatrix}},\;\mathbf{X} = \trsp{\begin{bmatrix} x_1 & \dots & x_M \end{bmatrix}}$, and define the data set $\cD_M=\{\mathbf{Y},\mathbf{X}\}$. GPR assumes that $y_i\sim\cN(d(x_i),\sigma_n^2)$ and $d\sim\cN(0,K_{d}(x,x'))$ for a kernel $K_d$. The choice of kernel functions depends on the particulars of the problem. In this paper, we use the Mat\'ern Kernel \cite{genton2001classes}. We can then define posterior distributions at any point $x\st\in\R$ given $\cD_M$ as $d(x\st)|\mathbf{Y}\sim\cN(\mu_d(x\st),\Sigma_d(x\st))$. The mean and variance functions $\mu_d(x\st)$, $\Sigma_d(x\st)$ of the GP model are defined as
\begin{align*}
    \mu_d(x\st) &= \trsp{\bK\st(x\st)}\inv{(\bK+\sigma_n^2)}\mathbf{Y},\\
    \Sigma_d(x\st) &= \bK\sst(x\st) - \trsp{\bK\st(x\st)}\inv{(\bK+\sigma_n^2)}\bK\st)(x\st).
\end{align*}
The terms $\bK\sst(x\st)$, $\bK\st(x\st)$ and $\bK$ are defined based on the kernel $K_d$ of the GP model: $\bK\sst(x\st) = K_d(x\st,x\st)\in\R$, $\bK\st(x\st) = K_d(\mathbf{X},x\st)\in\R^M$, $\bK = K_d(\mathbf{X},\mathbf{X})\in\R^{M\times M}$. Extensive further reading on this topic can be found in \cite{williams2006gaussian,bishop2006pattern}. One of the main benefits towards the risk-averse efforts in this paper of using GPR is that estimates are computed in predictive distributions. The resulting distributions will provide tools for risk assessment in Sections \ref{sec:ce} and \ref{sec:almain}. For analysis purposes we also define $\mathcal{O}=\{\varpi\in\R:\min_ix_i\leq\varpi\leq \max_ix_i\}$. The set $\mathcal{O}$ is the observed set that tracks which points were measured in the domain of $d$.

\begin{figure}
      \centering
      \includegraphics[trim=0mm 0mm 0mm 0mm, clip, width=0.30\textwidth]{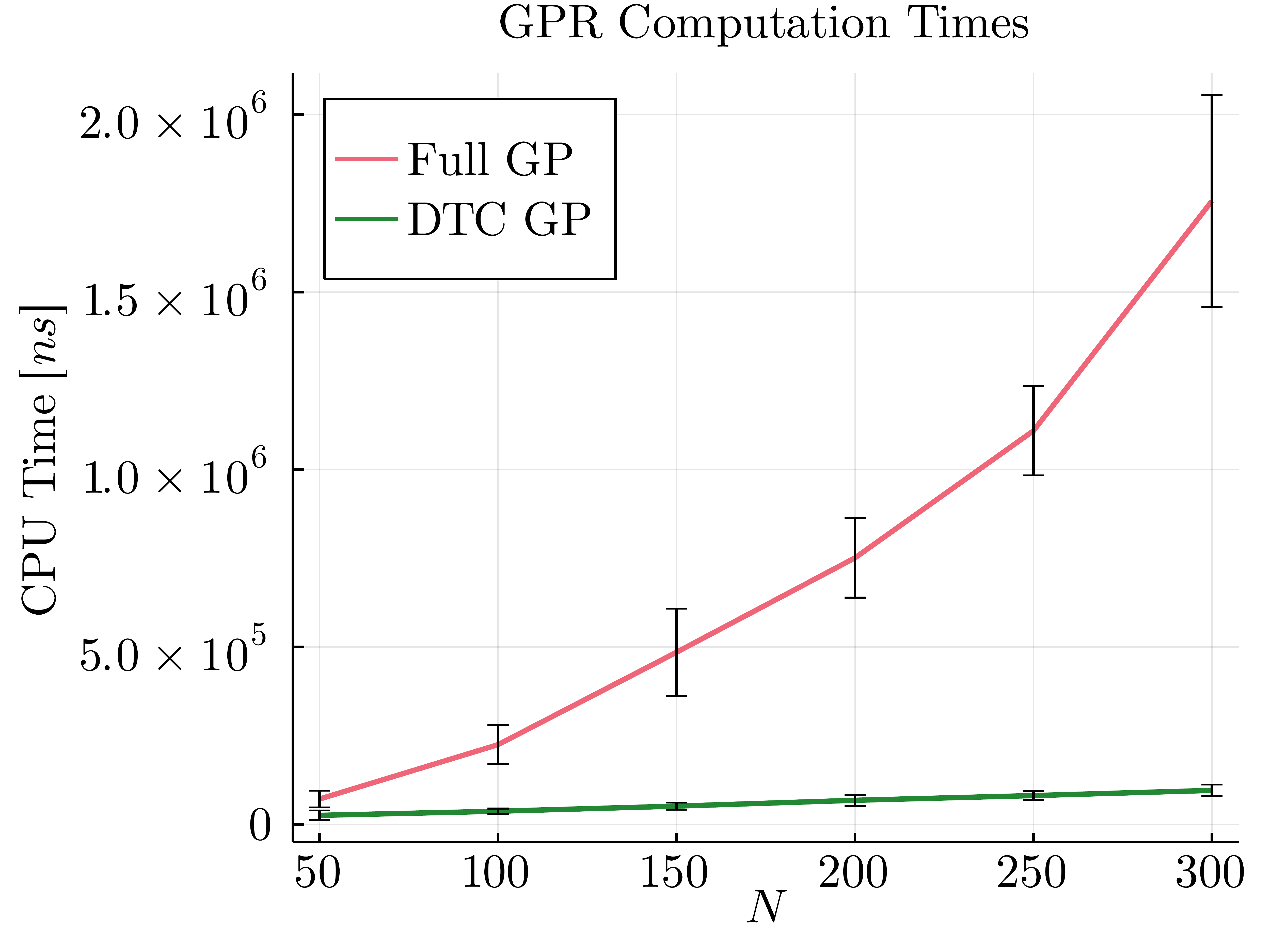}
      \caption{Median computation times for full GPR and DTC GPR. Bars represent standard deviation, $N$ indicates the amount of data points. At $N=300$ a full GP regresses in a median time of $1.555\si{ms}$, while DTC finishes in $88.827\si{\micro\second}$. All computations done on a single core of a 2012 Intel Core i7.}
      \label{fig:gprtimes}
   \end{figure}

\begin{figure}
      \centering
      \includegraphics[trim=0mm 0mm 0mm 0mm, clip, width=0.30\textwidth]{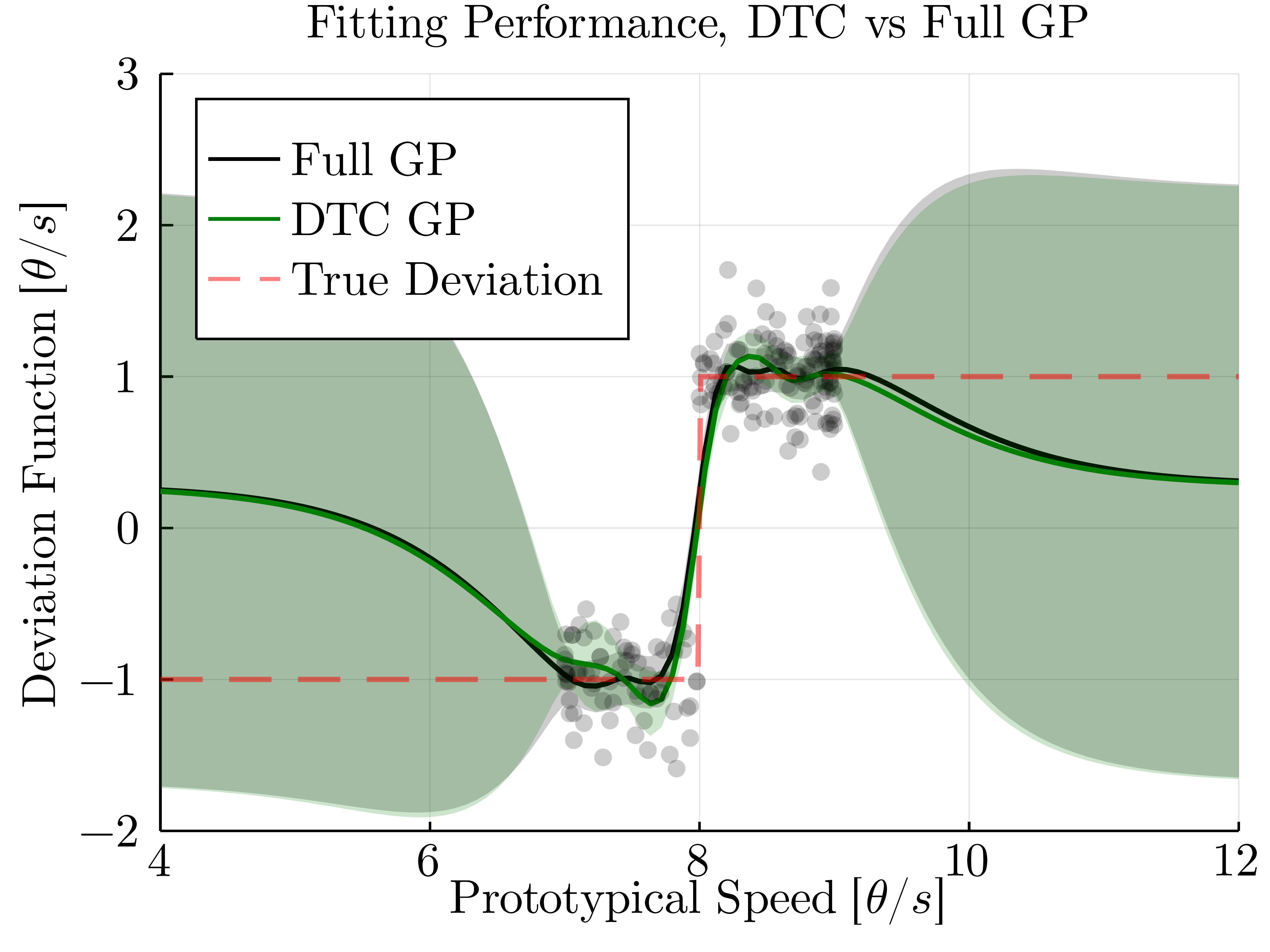}
      \caption{Fitting performance: DTC performs effectively the same as a full GP in this application. The goal is to approximate the true deviation function from observed data. Shaded area indicates 95\% confidence bounds.}
      \label{fig:gprcomp}
   \end{figure}

One downside of using GPR is computational efficiency. In this paper, we mitigate this issue with Sparse Gaussian Processes \cite[Sec.~8.4]{williams2006gaussian}. In particular, we use Deterministic Training Conditionals (DTC) \cite{liu2020gaussian,csato2002sparse}. Although DTC is not state-of-the-art, in practice and for this problem, in particular, it is not outperformed by other methods while providing non-negligible speedup. Deterministic Training Conditionals work by selecting specific inducing points instead of regressing over the entire data set. There are many approaches for selecting the inducing points; we use equally-spaced data quantiles. Compared to other methods, DTC had the property of being conservative --- an important characteristic for this framework. Figure \ref{fig:gprtimes} compares DTC to full GPR. Figure \ref{fig:gprcomp} shows a fitting performance comparison between the two methods.

\subsection{MPC formulation}
\label{sec:mpc}

In this section, we discuss the structure and particulars of the MPC component. A primary challenge of the rendezvous problem is presented by strict and numerous constraints, of which many are non-convex. By exploring two unique features of the problem formulation, we reduce dimensionality and attain tractability. We now outline the Optimal Control Problem (OCP) associated with the rendezvous problem. As mentioned previously in Sec.~\ref{sec:problem}, the solver is tasked with computing the Point-of-No-Return (PNR) and control inputs that navigate the UAS between important waypoints (rendezvous location, landing location). To fully define the problem and gain temporal constraint management, we expand the control from velocities to include a time ``input''. The nature of this problem requires the UAS to coincide with the vehicle both in space and time. By introducing time as a manipulated variable in the OCP, we allow the solver to decide on the optimal time allotment before reaching PNR directly. This feature is critical because we rely on maximizing this time allotment (decision time) to increase the number of data points we can gather and subsequently improve the GPR model's quality. This time input works by assuming a piecewise constant control law along each of the four segments (PNR, rendezvous, landing location, and abort location), which is possible due to our assumption on the UAS integrator dynamics described in \eqref{eq:sysdyn} and \eqref{eq:edyn}.

We represent each of the segments using the state vector $(\bx,\bv,\bt)\equiv(x_i,v_i,t_i)$, $i\in\{1,\dots,4\}$. Here, $t_i$ represents the time to be spent at a constant velocity $v_i$ to reach one waypoint from another. Furthermore, $x_i\in\R^2$ represents each of the defined physical waypoints in Euclidean coordinates and $v_i\in\R^2$ represents velocity inputs in Euclidean coordinates. The waypoints are, in this order, the Point-of-no-Return (PNR), the rendezvous location (RDV), the landing location ($S_L$), and the abort location ($S_A$), as shown in Figure \ref{fig:overview}. The designed Optimal Control Problem (OCP) is given by:
\begin{subequations}\label{eq:ocp}
\begin{alignat*}{3}
    \min_{U}\quad       & t_2+t_3+t_4-t_1    && \\
    \text{s.t.}\quad    & x_i = x_{i-1} + v_it_i,~x_4 = x_1 + v_4t_4,  && \\
                        & E_{r,k} = E_{r,k-1} - \left(\dfrac{m_kv^2}{2} + \alpha m_k\right) t_k, &&\\
                        & E_{r,4} = E_{r,1} - \left(\dfrac{m_4v^2}{2} + \alpha m_4\right) t_4, &&\\
                        & |v_i| \leq v_\mathrm{max},~ x_3 = p\st,~ x_4 = S_L,~x_5 = S_A,   && \\
                        & \sum_{i=1}^3 t_i \leq t_\mathrm{max},~t_1 + t_4 \leq t_\text{max},~t_c \leq t_i&& \\
                        & E_1 + E_2 + E_3 \leq E_{r,k},~E_1 + E_4 \leq E_{r,k}, 
\end{alignat*}
\end{subequations}
where $t_R \equiv t_1+t_2$ is a provided rendezvous time, $m_1=m_2=m_4=m_a$ is the UAS mass with the package, $m_3=m_b$ is the UAS mass without the package, $p_\mathrm{tgt}\in\cP$ is target path we aim to rendezvous with, $p\st$ is the optimal rendezvous location provided by the GPR model in Section \ref{sec:gpd} and elite samples from the importance sampling algorithm in Section \ref{sec:ce}, $S_L$ and $S_A$ are the landing and abort destinations, $E_{r,0}$ is the remaining energy, $E_{r,i},\;i=1,\dots,4$ is the energy associated with the segment, and $t_c$ is a dwell time for the low level controller to switch tracked segments. The dwell time is necessary to stop the solver from placing waypoints arbitrarily close to each other and creating undesirable sharp turns, which are problematic for our single integrator dynamics assumption. Moreover, $U\equiv\{t_i,v_i\},\;i\in(1,...,4)$. \textcolor{black}{Note that apart from risk-related design choices, all constraints and constants are given a priori from mission parameters.}
%All of these constraints are natural because every single one is predetermined at the design stage. For example, constraint \eqref{eq:er1} is directly produced from the battery or fuel tank used in the UAS, and constraint \eqref{eq:abp} is given from the map where the mission takes place. The only task left for the designer is to choose the risk measures in the importance sampling algorithm presented in Section \ref{sec:ce}.
\subsection{Importance Sampling}
\label{sec:ce}
In this section, we discuss the Sampling-based optimization algorithm. The goal is to approximate $\cA\st$, an $N$-dimensional distribution of optimal rendezvous times according to some criteria. In possession of a rendezvous time, we use the model found in Subsection \ref{sec:gpd} to estimate where the rendezvous location is. We perform this optimization problem using the cross-entropy (CE) algorithm \cite{botevCE}. We now set up this optimization problem. We present this algorithm in three parts; first, we discuss the ranking system that selects the best samples in a group; second, we add a risk-averse component; and finally, we briefly show how to update the sampling parameters.

\subsubsection{Sample Ranking}
Let $\cA(\mu_\cA,\Sigma_\cA)$ be a $N$-dimensional Gaussian distribution of rendezvous times with mean vector $\mu_\cA$ and diagonal variance matrix $\Sigma_\cA$. Let $n_s$ and $n_e$ be positive integers such that $n_s>>n_e$. Let $\cS\in\R^{N\times n_s}$ be a matrix of $n_s$ samples from $\cA$. We say that $\cS$ is the sample set, and $\cS_e\in\R^{N\times n_e}$ is the elite sample set that contains the $n_e$ best row-wise samples from $\cS$. In other words, $\cS_e$ is a matrix where each row contains the $n_e$ best samples for the path associated with that row. To find the elite set, we partially rank the sample set according to a a cost function $l(n^{i,j})$ we wish to minimize, where $n^{i,j}$ is the element in the $i$-th row and $j$-th column of the (non-ordered) set $\cS$. Each element $n^{i,j}$ is a time sample that produces a rendezvous location candidate. The first step is to compute the expected driver position for each sample as a rendezvous location. Consider a moment in time $t_0$ and a time sample $t^{i,j}\in\cS$. At $t_0$ we have a GPR model of $d(\dot{\theta}^h_i(t))$ for every path $i\in\cP$. Thus
\begin{align*}
\E[\theta^d_i(t)] %&= \int_{t_0}^{t^{i,j}}\dot{\theta}^h_i(t) + d(\dot{\theta}^h_i(t))\diff{t} \\
= \theta^d(t_0) + \int_{t_0}^{t^{i,j}}\dot{\theta}^h_i(t) + d(\dot{\theta}^h_i(t))\diff{t},\numberthis\label{eq:driverposition}
\end{align*}
where the integration procedure is done numerically using Gauss-Kronrod Quadrature. This integration completes in a median time of $124.4\si{\micro\second}$ on a single core of a 2012 Intel Core i7. However, the nature of the algorithm permits this implementation to be largely computed in parallel on a GPU, although such implementation is not done in this paper. A parallel implementation would provide significant speedup for higher values of $n_s$ than the ones used in this paper (we use $n_s=5$, for reference). Using \eqref{eq:driverposition} we can compute the expected driver position $\bp$ for each time sample $j$ and path $i$ with $p_i(\E[\theta^d_i(t_j)])$ in matrix form:
\begin{align*}
    \mathbf{p} = \begin{bmatrix}    p_1(\E[\theta^d_1(t_1)]) && \cdots && p_1(\E[\theta^d_1(t_{n_s})]) \\
                                    \vdots && && \vdots\\
                                    p_N(\E[\theta^d_N(t_1)]) && \cdots && p_N(\E[\theta^d_N(t_{n_s})])
    \end{bmatrix}.
\end{align*}
In possession of $\bp$ we can compute the ``quality'' of each sample. Naturally, because our goal is to not run out of fuel or  battery, we choose samples that minimize energy consumption. This is not equivalent to minimizing distance to the UAS for two reasons: (a) spatial points have a temporal constraint, and (b) the landing location and remaining fuel for landing depend on an external path planner (discussed in Subsection \ref{sec:mpc}). Temporal constraints mean that two equally close rendezvous candidates have two different times for the UAS to reach that location. The energy dynamics \eqref{eq:edyn} are such that the best rendezvous time is non-obvious in this case. Additionally, the landing location and remaining fuel after rendezvous need to be included in the quality criteria. Failure to do so would select a sample that minimizes the energy necessary to rendezvous, but might not minimize overall mission energy. To compute energy costs for each sample we use the energy dynamics \eqref{eq:edyn}. Let $\mathbf{E}\in\R^{N\times n_s}$ be a matrix containing the energy costs for each sample, and $x_0=[x_0^1,x_0^2]$ be the euclidean position of the UAS at $t_0$. Then we compute each element of $\mathbf{E}$ as
\begin{align*}
    \mathbf{E}^{i,j} = &m_a(t_j-t_0)\left[\frac{1}{2}\tnorm{v_r^{i,j}}^2+\alpha\right] +\\ &m_b(t_l-t_j)\left[\frac{1}{2}\tnorm{v_l^{i,j}}^2+\alpha\right],\numberthis\label{eq:eterms}
\end{align*}
where $t_l=\sum_{k=1}^3t_k$ is provided by the MPC solution in Section \ref{sec:mpc} (in the absence of a solution in the first iteration we do not compute the second term of \eqref{eq:eterms}), $\tnorm{\cdot}$ denotes the 2-norm of a vector, and
\begin{align}
    v_r^{i,j} &= \frac{|\mathbf{p}^{i,j}-x_0|}{t_j-t_0},\quad v_l^{i,j} = \frac{|S_L-\mathbf{p}^{i,j}|}{t_l-t_j},\label{eq:vneutral}
\end{align}
where $S_L$ is the landing location as described in Section \ref{sec:mpc}. For this paper, we set $l(n^{i,j}) = \mathbf{E}^{i,j}$. In Subsection \ref{sec:riska} we add a risk-averse component to the calculation of $\mathbf{E}$. These equations reflect the (generic) energy dynamics we consider in this paper, but the procedure is agnostic to the energy dynamics chosen by the designer. To find $\cS_e$ we select the $n_e$ best samples for each path:
\begin{align*}
    \cS_e = \begin{bmatrix} \argmin_j^{n_e} l(n^{1,j}) \\
    \vdots \\
    \argmin_j^{n_e} l(n^{N,j})
    \end{bmatrix},
\end{align*}
where $\argmin_x^{k}f(x,\dots)$ means we select the $k$-best arguments that minimize $f$ over $x$ that we output in the order from best to worst. This way the first column of $\cS_e$ contains the best sample for each path and so on for the other columns. Using similar logic, we select the target optimal rendezvous location $p\st = p_\mathrm{tgt}(\theta^d_\mathrm{tgt}(t_{R,\mathrm{tgt}}))$ to be forwarded to the MPC planner. Let $S_e^{i,1}$ be the first column of $S_e$. Then for $\mathrm{tgt}\in\{1,\dots,N\}$ let $\mathrm{tgt} = \argmin_i c\left(\cS_e^{i,1}\right)$. This equation applies a cost function to the best samples from each path and selects the best path index based on that cost function. The design of $c$ is intricate, and an in-depth analysis of what constitutes a suitable cost function is left as future work. We present two naive designs in this paper.

The most natural function one could choose is based on the \emph{Best First} logic. This logic selects the best global sample, and we consider the path associated with that sample to be the optimal one to rendezvous with. This is equivalent to simply setting $c(\cS_e^{i,1}) = \cS_e^{i,1}$, i.e. select the time sample of least energy. This strategy would be desirable if it is possible to control which path the driver will choose. In a different framework application where we would consider autonomous vehicles as the ground agent, such a cost function is highly desirable. In this paper, however, selecting the best global sample is overly ambitious. If the driver chooses any path other than $p_\mathrm{tgt}$, there is a non-negligible probability that there will not be enough battery to alter the course since the MPC will be spending resources to maximize $t_1$. An alternative version of Best First applies weights to the cost function with $c(\cS_e^{i,1}) = w_i\cS_e^{i,1}$,
where $w_i\in\R^N_+$ is a vector of user-defined weights. This variant is beneficial if the designer has prior knowledge of the driver's probability of choosing each path.

In this paper, we use the opposite strategy; \emph{Worst First}. This strategy selects the worst-of-the-best time samples, i.e., from the best samples for each path, selects the worst path. The logic is simple: if Best First is an ambitious strategy, Worst First is a conservative one. If we plan to have enough energy to reach the worst path for optimal rendezvous, all others require less energy and, thus, are reachable. This strategy assumes $c(\cS_e^{i,1}) = -w_i\cS_e^{i,1}$, where we can again weigh each entry according to some path choice distribution. For the rest of this paper, we assume a uniform path choice distribution such that $w_i=1\forall i$.On an implementation note, we use the partial quicksort algorithm in \cite{martinez2004partial} to quickly partially sort the array of samples.
%Partial quicksort is more efficient than naive approaches since it only returns the top-$k$ elements of an array. The partial sorting procedure finishes on a median time of $2.182\si{\micro\second}$ on a single core of a 2012 Intel Core i7. Since all entries are independent, this procedure is both vectorized and parallelized so that CPU time scaling is better than linear.
\subsubsection{Risk Assessment}
\label{sec:riska}

Quantifying risk is the effort to determine a measure $\rho$ that maps a set of random variables to a real number representing the probability or expected value of an undesirable outcome \cite{riskmeasures,riskbook}. With this definition, the random variables are the states of the UAS (due to process and measurement noises) and, more importantly, the position of the ground vehicle due to the driver's uncertain behavior. It is crucial to choose measures that reflect meaningful quantities in the problem formulation. In this framework, risk directly relates to uncertainty regarding the vehicle's location in the future and the limitations that the path imposes on planning. If the driver is erratic, or the path only allows the rendezvous to happen in unfavorable locations, we consider that the mission has elevated risk. Several risk measures are popular; some examples are Expectation-Variance \cite{whittle}, (Conditional, Tail) Value-at-Risk ~\cite[Sec.~3.3]{riskbook}, and Downside Variance ~\cite[Sec.~3.2.7]{riskbook}. These measures can introduce nonlinearity and preclude gradient information, endangering tractability. A popular approach uses gradient-free methods, which sample these measures and choose inputs corresponding to minimum risk \cite{samprisk}. In this paper, we use two risk measures; $\rho_r$ is the rendezvous risk measure used by the rating system to select samples of least risk, and $\rho_d$ is the decision risk measure used in Subsection \ref{sec:almain} to decide on whether to abort the mission and safely return, or proceed with the rendezvous. In this subsection, we discuss the design and implementation of $\rho_r$. The main differentiator between the two is that the rendezvous risk measure needs to be computationally efficient since we repeat its calculations for every sample, every time step. We embed risk into the cost by adjusting the distance between each sample and the UAS or landing area in the numerator of Eq.~\eqref{eq:vneutral}.

We start by computing the propagated uncertainty for each sample $n^{i,j}$ as
\begin{align}
    h^{i,j} = \gamma\int_{t_0}^{n^{i,j}}\Sigma_{d,i}(\dot{\theta}^h_i(t))\diff{t}\label{eq:uint},
\end{align}
where $\gamma\in\R^+$ is a scaling factor. Equation \eqref{eq:uint} makes it so that $p_i(\E[\theta^d_i(t_j)])\pm h^{i,j}$ represents a confidence interval in $\theta$. The goal now is to select which of these three numbers is furthest from the UAS or landing location, and use that distance when computing necessary velocities. Let $\dis(a,b):\R^2\times\R^2\mapsto\R$ be the Euclidean distance between two points $a,b\in\cR$, then for every sample $n^{i,j}$ let $r^{i,j} = \dis(\mathbf{p}^{i,j},x_0)$ be the neutral range, $r^{i,j}_+ = \dis(p_i(\E[\theta^d_i(t_j)])+h^{i,j},x_0)$ be the positive uncertainty range, and $r^{i,j}_- = \dis(p_i(\E[\theta^d_i(t_j)])-h^{i,j},x_0)$ be the negative uncertainty range. Then $\rho_r$ naturally follows: $\rho_r^{i,j}(\Sigma_d,\cS,x_0) = \max(r^{i,j},r^{i,j}_+,r^{i,j}_-) - r^{i,j}$.

\begin{figure}
    \centering
    \includegraphics[trim=170mm 160mm 270mm 110mm, clip, width=0.3\textwidth]{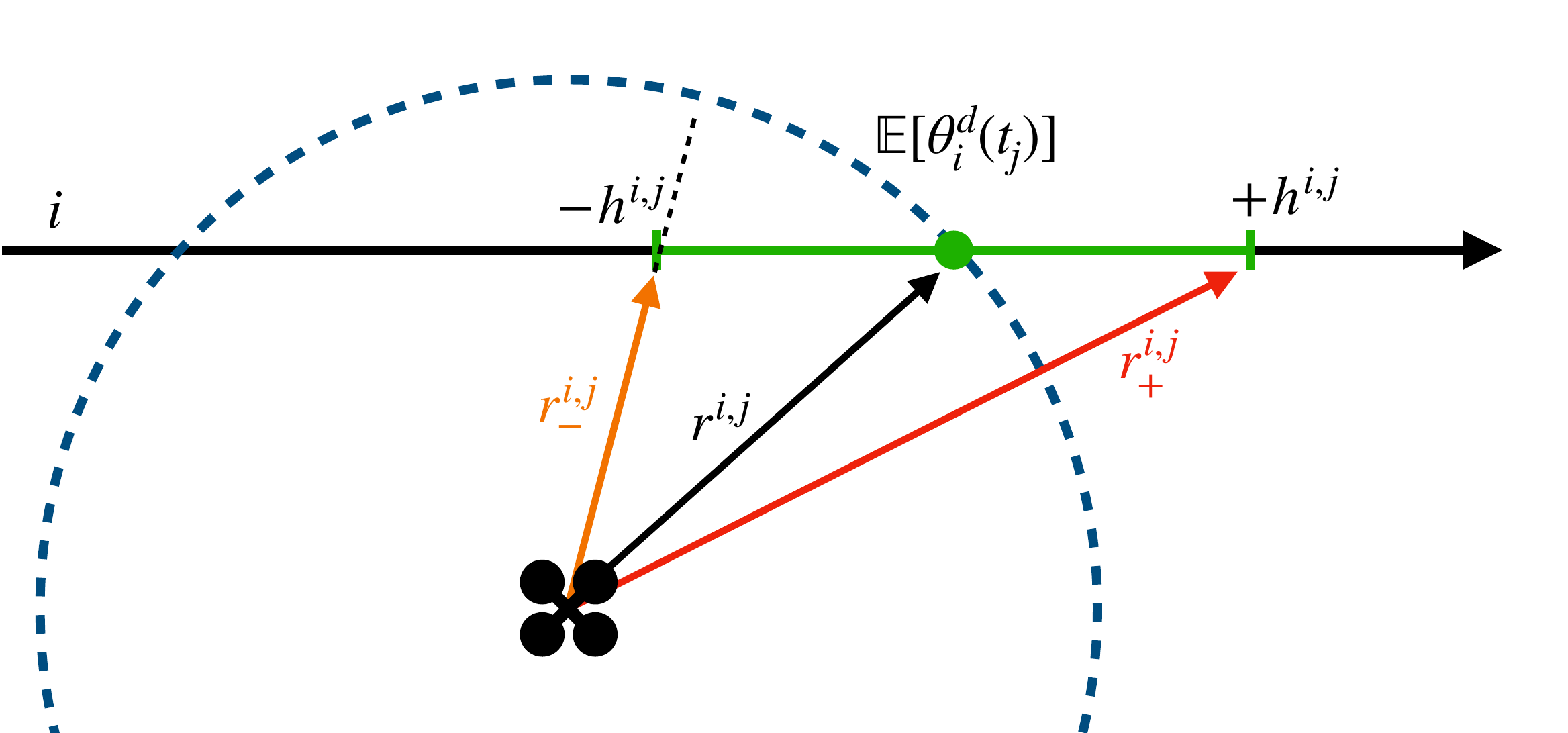}
    \caption{Downside Risk as potential required range gain. The red outcome forces the UAS to spend more energy to meet the car. The extra energy is the downside potential, used as risk measure.}
    \label{fig:dsr}
\end{figure}

Figure \ref{fig:dsr} shows a visual representation of the different ranges. This risk measure is an approximation of Conditional Value-at-Risk ($\Cvar$). In its common form, $\Cvar_\gamma$ represents the expected value of the $\gamma$-percentile of a distribution that quantifies potential loss (downside potential). Here, instead of computing the energy distribution, we compare the energy required to reach the sample at its mean and at some $\sigma$-distance away from the mean. We then pick the worst outcome and say this is the potential loss for this sample. Finally, we can compute the risk-enabled velocities with
\begin{align*}
    v_r^{i,j} &= \frac{r^{i,j}+\rho_r^{i,j}(\Sigma_d,\cS,x_0)}{t_j-t_0},\\ 
    v_l^{i,j} &= \frac{r^{i,j}+\rho_r^{i,j}(\Sigma_d,\cS,S_L)}{t_l-t_j},
\end{align*}
and compute and rank $\mathbf{E}$ the same way as before.

\subsubsection{Parameter Update}
\label{sec:paramup}
This section discusses the parameter update algorithm for a single path. Since all paths are independent, we repeat this process identically for every path. We update $\mu$ and $\Sigma$ by taking mean and variance row-wise from $\cS_e$ with $\mu_\cA = \mathrm{Mean}(\cS_e)$ and $\Sigma_\cA = \Var(\cS_e) + \lambda$, where $\lambda\in\R$ is a small positive number. The scalar parameter $\lambda$ serves as an exploration tool due to the time-varying nature of the algorithm, and avoids convergence to a (traditionally desirable) static impulse-like distribution. 

\subsection{Heuristics}
\label{sec:almain}
In this section, we discuss the architecture of the overarching algorithm that integrates all modules and commands a rendezvous/abort decision. In summa, this is shown in Algorithm \ref{al:main}. The functions in Algorithm \ref{al:main} and their correlated method are shown in Table \ref{tab:funs}.
\begin{algorithm}[ht]
\SetAlgoLined
$\mathcal{D}\leftarrow$ Initial Data\\
 \While{$t_1>\epsilon$}{
    $d,\;\Sigma_d \leftarrow \text{Regress}(\mathcal{D})$\\
    $\cS \leftarrow \text{Sample}(\cA,N_s)$\\
    $p\st,\;\cS_e \leftarrow \text{Rank}(\cS,\bx,\bt,d,\Sigma_d)$\\
    $\mu_\cA,\;\Sigma_\cA \leftarrow \text{UpdateParameter}(\cS_e)$\\
 	$\bv,\bx,\bt\leftarrow \text{MPC}(p\st,\bx)$\\
 	Send Control Input $\bv$ to UAS\\
 	$\mathcal{D}\leftarrow \text{Append(Sensor Data},\mathcal{D})$}
  \uIf{$\rho_d(d,\Sigma_d,\bx)\leq \kappa$}{
    Proceed with rendezvous and then to $S_L$
  }
  \Else{
    Abort and return to $S_L$
  }
 \caption{Mission Algorithm}
 \label{al:main}
\end{algorithm}
\begin{table}[ht]
    \centering
\begin{tabular}{l|l}
Function Name   & Procedure                             \\ \hline
Regress         & GPR in Sec.~\ref{sec:gpd}                       \\
Sample          & Returns $N_s$ samples from $\cA$      \\
Rank            & Ranking Procedure in Sec.~\ref{sec:ce}             \\
UpdateParemeter & Updates parameters of $\cA$ as in Sec.~\ref{sec:paramup} \\
MPC             & Computes MPC control inputs as in Sec.~\ref{sec:mpc} \\
Append          & Appends new sensor data to $\cD$     
\end{tabular}
    \caption{Correlation between Algorithm \ref{al:main} and this paper's methods.}
    \label{tab:funs}
\end{table}

The overarching logic is the same as discussed in Section \ref{sec:intro}. While the decision time $t_1$ (time before reaching the PNR waypoint) is greater than some constant, we keep acquiring data, improving the model, and searching for a better rendezvous point. When a decision is necessary we perform a one-time risk analysis and comparison against the scalar constant $\kappa\in\R$ to decide between turning back or continuing. Note that following Definition \ref{def:persafe}, $\bx_k\in\cS_k\forall k$ because if this condition is violated, we switch to the plan with deterministic guaranteed safety.

\textcolor{black}{As discussed in Section \ref{sec:ce}, for online optimization purposes we approximate $\Cvar_\gamma$. When deciding whether to abort or attempt a rendezvous, however, we can afford an expensive one-time computation of $\Cvar_\gamma$. We briefly define $\Cvar_\gamma$ for completion. Let $\bZ$ be a set of real-valued continuous random variables, $\rho:\bZ\mapsto\R$ be a risk measure function, $\bX\in\bZ$ be a random variable, $x\in\Omega_\bX$ be the domain of $\bX$, $f_\bX(x)$ be the probability density function of $\bX$, $F_\bX(x)$ be the cumulative density function of $\bX$. Then we define the two quantities of interest: $\mathrm{VaR}_\gamma(\bX) = \inf\{x\in\Omega_\bX : F_\bX(x) \geq \gamma \}$ and
\begin{align*}
    \Cvar_\gamma(\bX) = \frac{1}{\gamma}\int_0^\gamma \mathrm{VaR}_\gamma(\bX) \diff{\gamma},
\end{align*}
where $\gamma\in[0,1]$ is a real-valued quantile. For the purposes of this paper, we consider the the distribution of extra fuel required for rendezvous ($\bX = E_e = E_r-(E_1+E_2+E_3)$) as the random variable, with distribution derived from $\mu_d$ and $\Sigma_d$ for each of the paths' optimal rendezvous location.}

\section{Results}
\label{sec:res}
\urlstyle{tt}
In this section, we present results on all modules. Implementation code that generates all figures and animations of results are available at \url{https://github.com/gbarsih/Multi-Path-Safe-Rendezvous}.

\subsection{Learning Performance}
\begin{figure}
      \centering
      \includegraphics[trim=0mm 0mm 0mm 0mm, clip, width=0.35\textwidth]{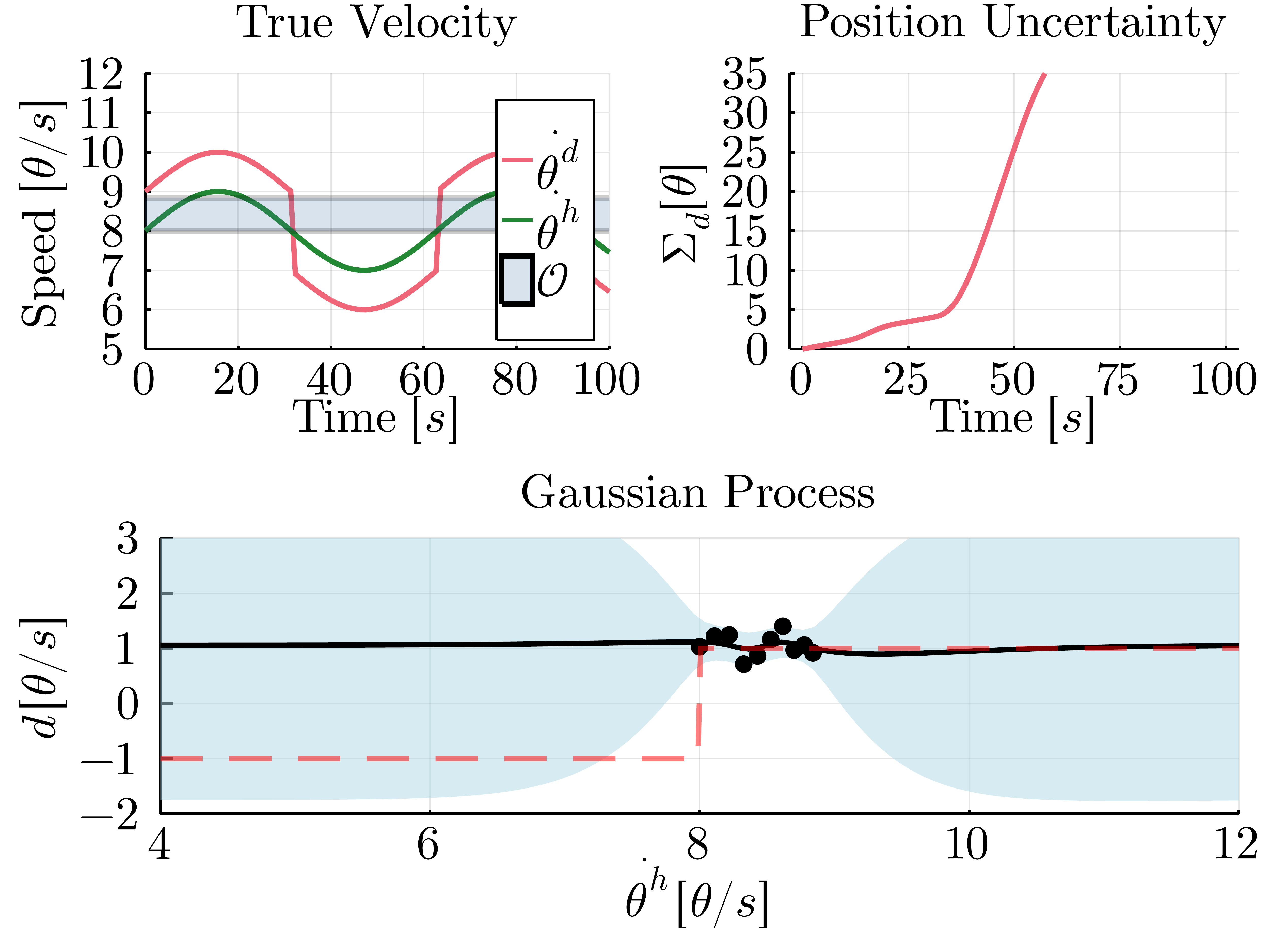}
      \caption{GPR learning process snapshot after $10\si{\second}$. Shaded area indicates 95\% confidence bounds.}
      \label{fig:learna}
   \end{figure}

\begin{figure}
      \centering
      \includegraphics[trim=0mm 0mm 0mm 0mm, clip, width=0.35\textwidth]{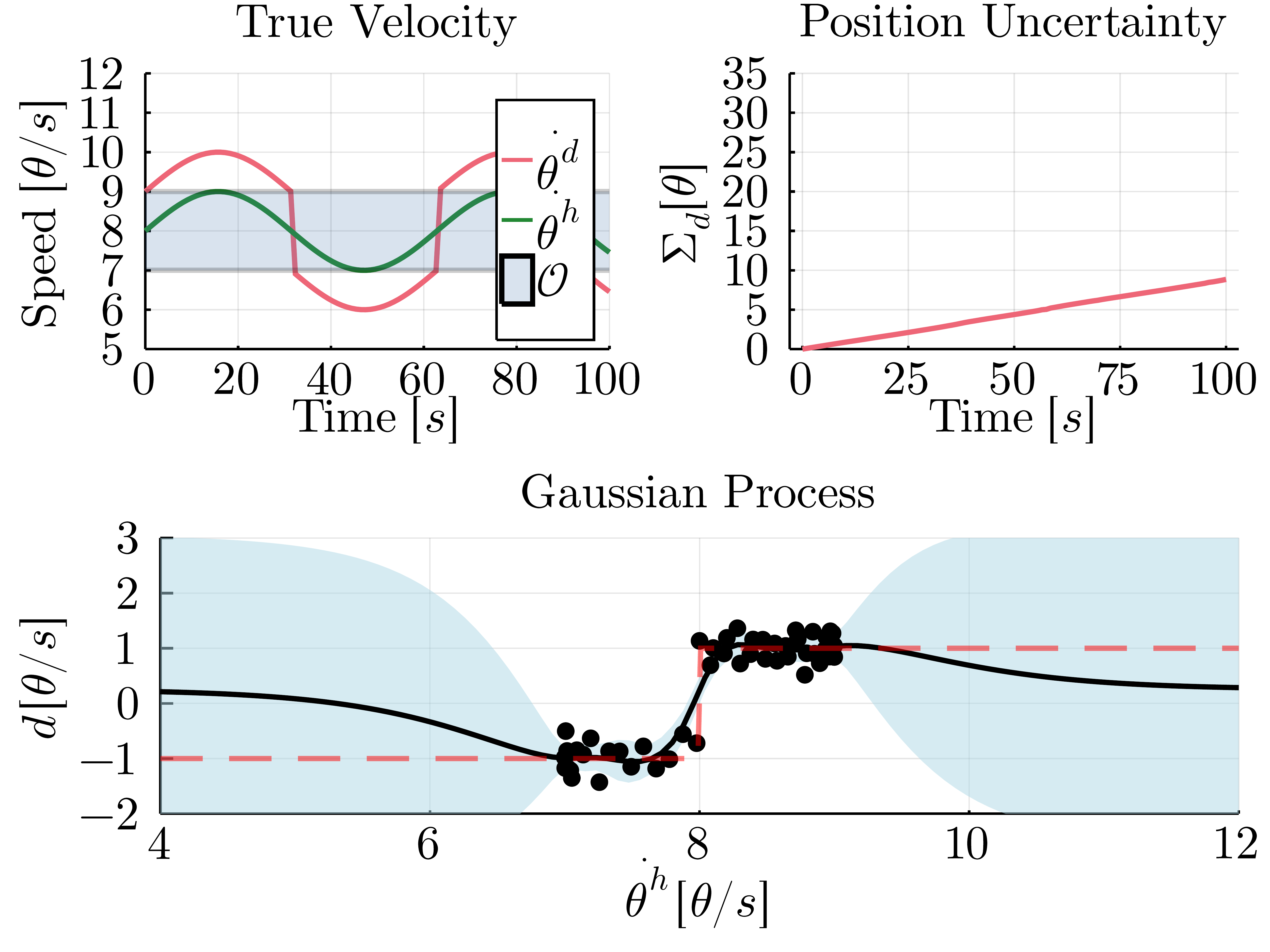}
      \caption{GPR learning process snapshot after $50\si{\second}$.}
      \label{fig:learnb}
   \end{figure}

We start by presenting results on the learning algorithm in Section \ref{sec:gpd}. The learning problem seeks to find a deviation function $d$ that captures the driver behavior. Figures \ref{fig:learna} and \ref{fig:learnb} show two points in time. Figure \ref{fig:learna} is taken after $10\si{\second}$ of data gathering, and Figure \ref{fig:learnb} after $50\si{\second}$. The nature of $\td^h$ ensures that in Figure \ref{fig:learna}, $\cO$ is limited to only under half of possible values $\td^h$ can take. Consequently, the GPR model has no information on that region and produces a high value for projected uncertainty. Conversely, in Figure \ref{fig:learnb} we explored the entire domain, and the GPR model can produce estimates with higher certainty. For the remaining results, we use the same functions shown here for all paths, where $\td^h(t)=8+\sin(t/10)$ and $\td^d(t) = \td^h(t) + \sign (\td^h(t) - 8)$.
\subsection{Importance Sampling}
\begin{figure}
      \centering
      \includegraphics[trim=0mm 0mm 0mm 0mm, clip, width=0.35\textwidth]{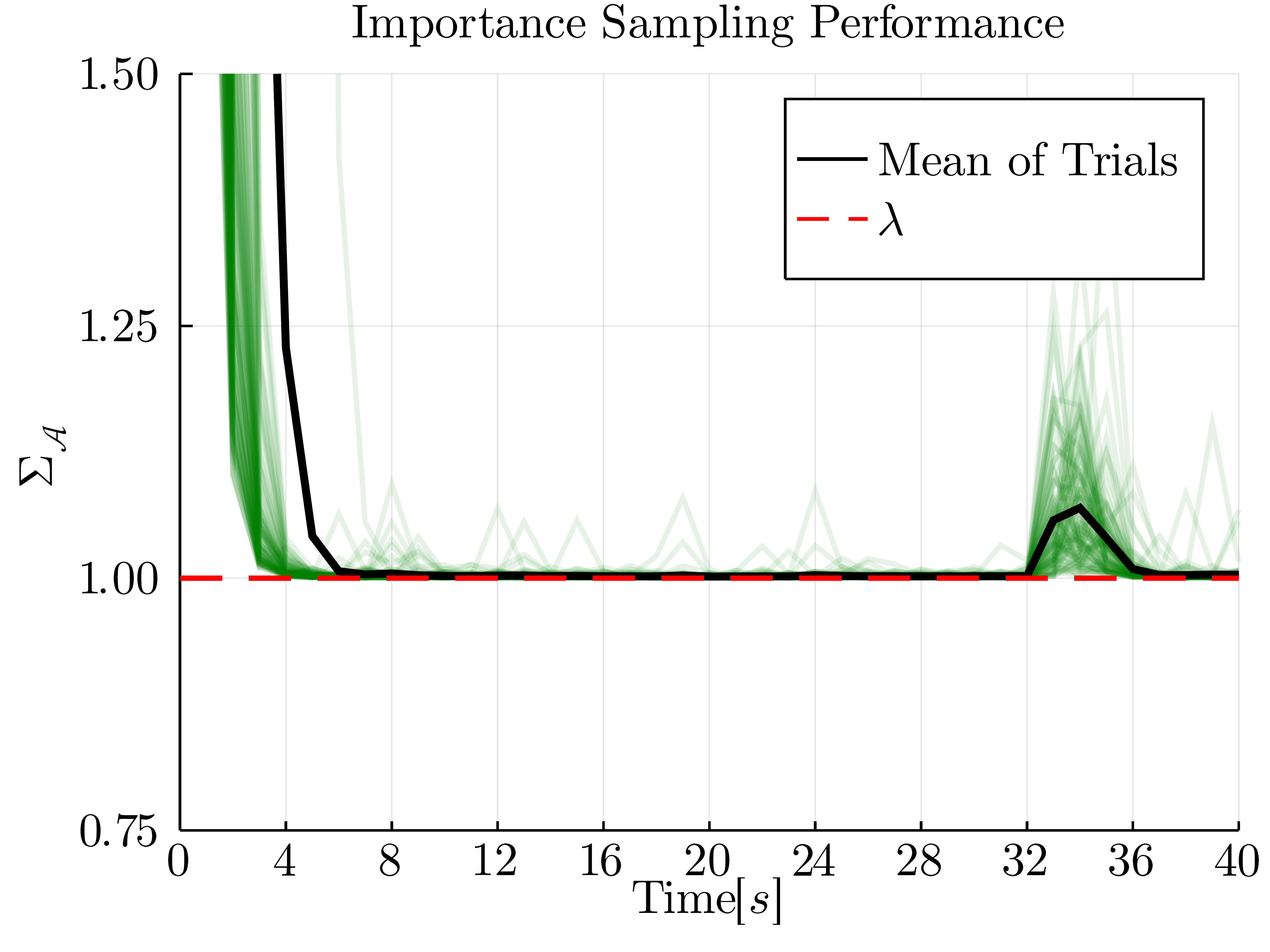}
      \caption{Average convergence rate (black) of $\Sigma_\cA$ for 100 trials (green). At $t=10\pi\si{\second}$, $\cO$ starts to cover new information. The shift in the learned driver behavior causes the sampling algorithm to react as indicated by the momentary increase in $\Sigma_\cA$.}
      \label{fig:ce_mdl}
\end{figure}

Here, we present results on the importance sampling algorithm in Section \ref{sec:ce}. Since an analytical form of the optimal rendezvous location does not exist, we leave the performance quantification for the full planning stack results in the next subsection. To show the efficacy of this individual module, we analyze the convergence rate. Figure \ref{fig:ce_mdl} shows 100 different runs of the same algorithm, with mission parameters randomly selected. We notice that it quickly converges (in about four iterations), and that when $\cO$ begins to expand at $t=10\pi$, the algorithm increases $\Sigma_\cA$ to optimize over the new data. 
%\begin{figure}
%      \centering
%      \includegraphics[trim=0mm 0mm 0mm 0mm, clip, width=0.4\textwidth]{ACC-Risk-2020/fig/ce_con.pdf}
%      \caption{Convergence rate .}
%      \label{fig:ce_con}
%\end{figure}

\subsection{Full Planning Stack}
\begin{figure}[h]
      \centering
      \includegraphics[trim=0mm 0mm 0mm 0mm, clip, width=0.35\textwidth]{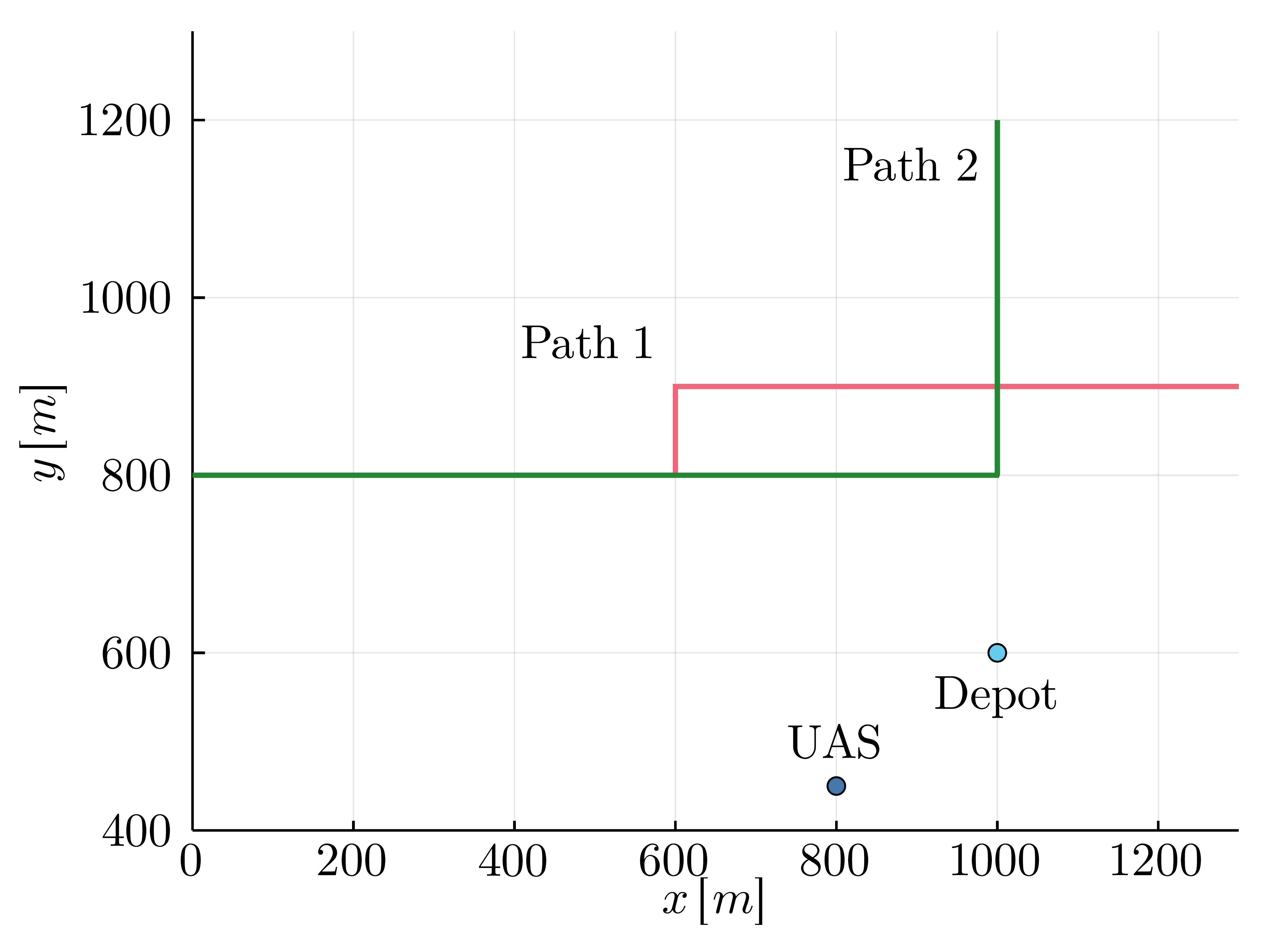}
      \caption{Mission map representing a section of an urban grid.}
      \label{fig:map}
\end{figure}

\begin{figure}[h]
      \centering
      \includegraphics[trim=0mm 0mm 0mm 0mm, clip, width=0.35\textwidth]{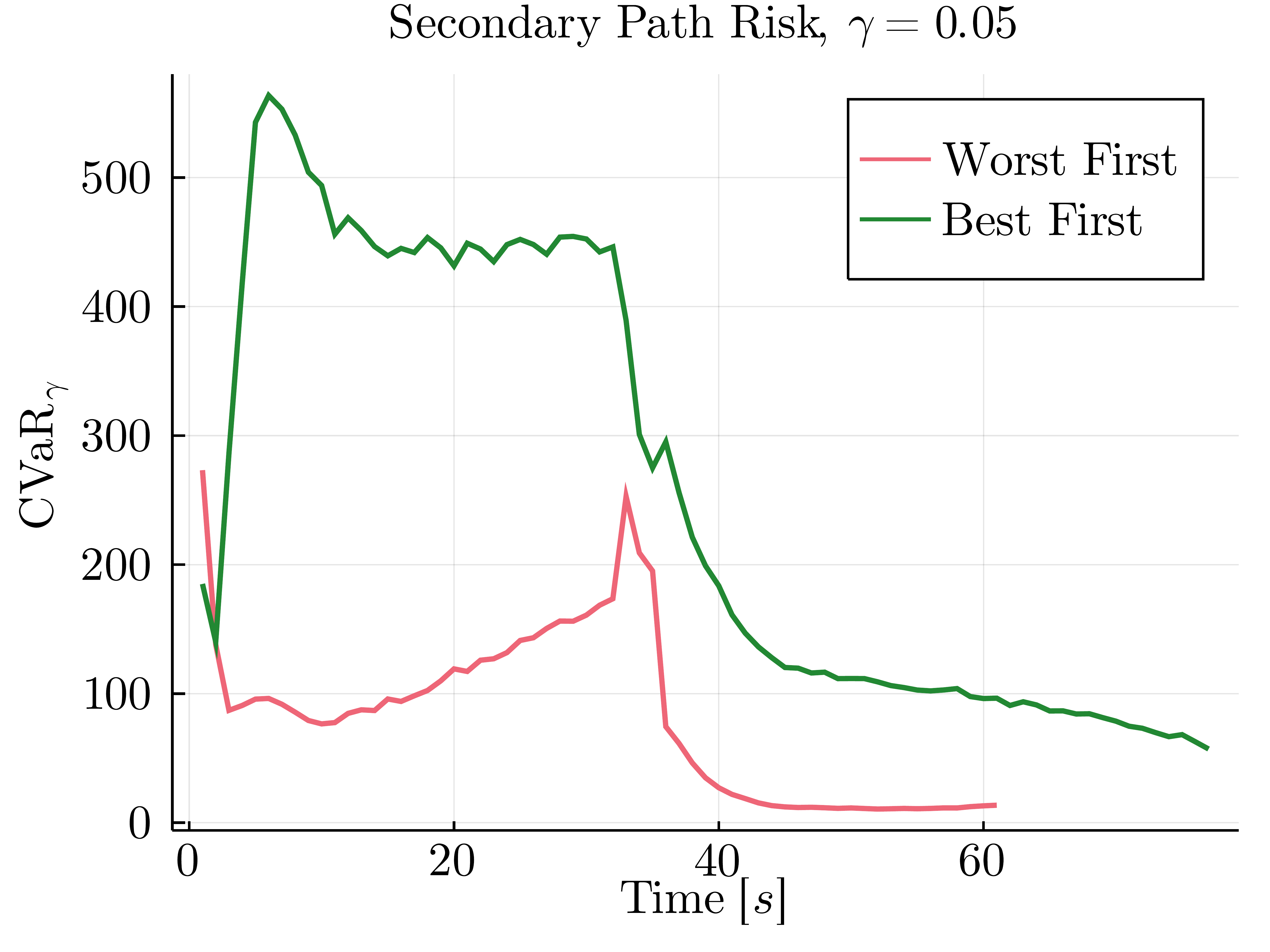}
      \caption{Same mission parameters, equally seeded, for the two strategies proposed in Section \ref{sec:ce}. Plot terminates when $t_1<\epsilon=1$. Worst First finds trajectories with least risk should the driver choose a path $p\in\cP\setminus p_\mathrm{tgt}$.}
      \label{fig:comprisk}
\end{figure}

\begin{figure}[h]
      \centering
      \includegraphics[trim=0mm 0mm 0mm 0mm, clip, width=0.35\textwidth]{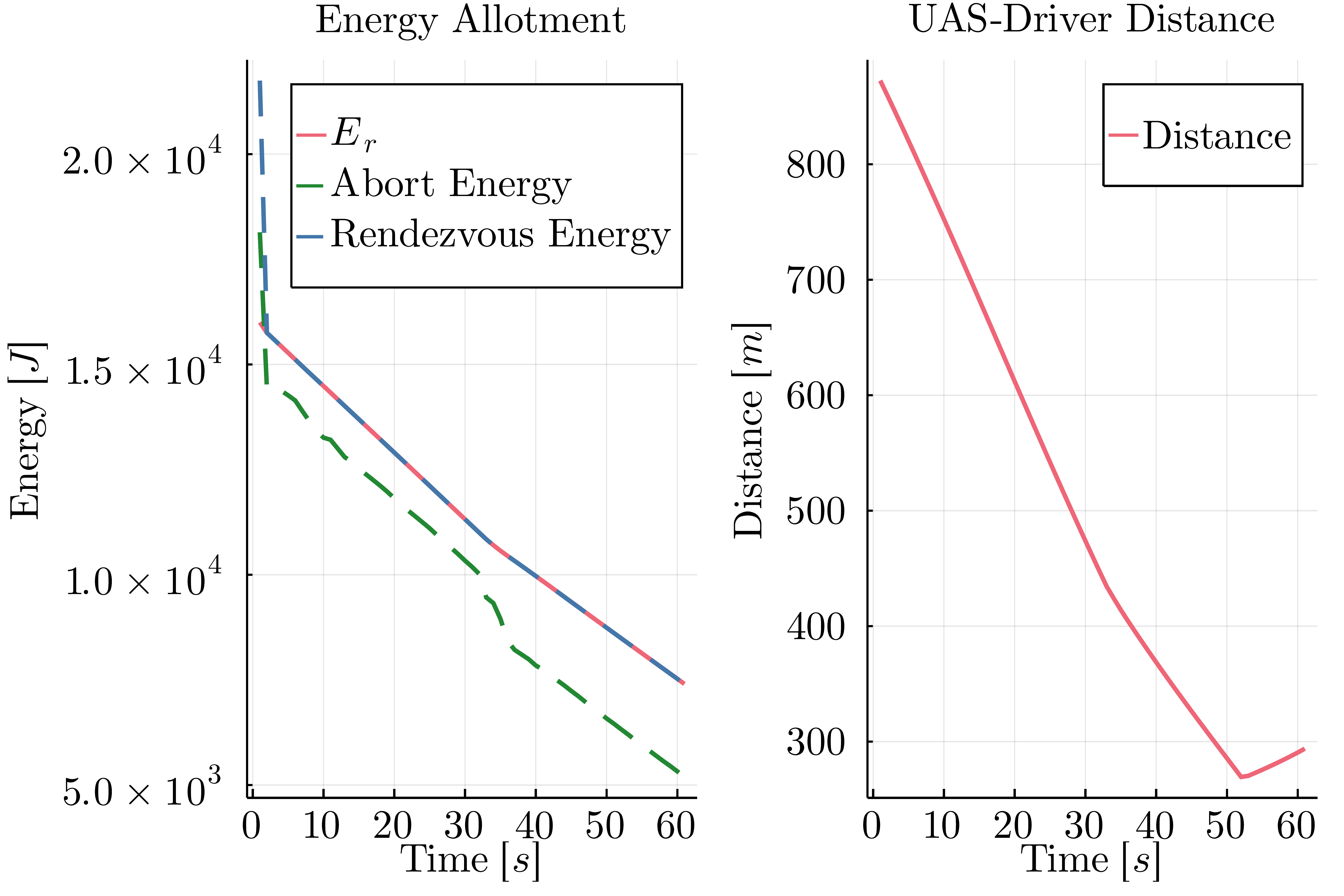}
      \caption{Results using Worst First strategy. The algorithm uses all available energy to try and minimize risk. Distance increase towards the end is due to path geometry.}
      \label{fig:energies}
\end{figure}

 Consider the map in Figure \ref{fig:map}.  With $m=[3,1]\si{\kilo\gram}$ the maximum range assuming no drop-off is $335\si{\metre}$, while the maximum range assuming a successful drop-off is $485\si{\metre}$, with $\alpha=20$ and $E_{r,0} = 1.6\mathrm{E}4$. Under these conditions, the roads are only reachable with the algorithm presented in this paper. Figure~\ref{fig:comprisk} compares the risk associated with the secondary path (that is, the path that is not $p_\mathrm{tgt}$) for the two importance sampling strategies. As expected, using Worst First yields reduced risk. The choice of $\kappa$ is entirely dependent on the mission parameters and how much risk the designer is willing to take; however, using the Worst First strategy will, in general, attempt a rendezvous more often. Note that which path the driver chooses after a decision is made, is irrelevant here since risk assessment is performed for all cases. The algorithm aims to ensure that risk is low for all possible outcomes. Finally, Figure \ref{fig:energies} depicts the planned energies throughout the mission and the distance from the UAS to the driver. Both plans (abort and rendezvous) are maximized and feasible.
 
\section{CONCLUSIONS}
\label{sec:conc}

We presented an algorithm capable of planning and executing a rendezvous mission between an autonomous UAS and a human-operated ground vehicle. The planner can assess the risk associated with the human factor and make informed decisions on either proceeding with the rendezvous or flying to a safe landing location. Such an arrangement is persistently safe because the abort plan is deterministic. We show numerically that the approach accomplishes its goals. For future work, we intend to address two deficiencies of this method. First, the algorithm needs to account for multiple drivers. There are untapped benefits of having multiple rendezvous options at any given time. Second, we wish to model the dynamics of drivers entering the network to preemptively start a mission and improve system efficiency.

%\section*{APPENDIX}

%%%%%%%%%%%%%%%%%%%%%%%%%%%%%%%%%%%%%%%%%%%%%%%%%%%%%%%%%%%%%%%%%%%%%%%%%%%%%%%%

\bibliographystyle{IEEEtran}
\bibliography{bib/bibliography.bib}

\end{document}